\documentclass[10pt,twocolumn,letterpaper]{article}

\usepackage{cvpr}
\usepackage{epsfig}
\usepackage{graphicx}
\usepackage{amsmath}
\usepackage{amssymb}
\usepackage{ulem}
\usepackage{placeins}
\usepackage{comment}
\usepackage{verbatim}
\usepackage{subcaption}
\usepackage{array}
\usepackage{booktabs}
\usepackage{float}
\usepackage{mathtools}
\usepackage{pdfpages}
\usepackage{times}
\usepackage{appendix}

\newcolumntype{M}[1]{>{\centering\arraybackslash}m{#1}}

\usepackage[pagebackref=true,breaklinks=true,letterpaper=true,colorlinks,bookmarks=false]{hyperref}

\cvprfinalcopy

\ifcvprfinal\fi

\newcommand{\fig}[1]{Fig.~\ref{fig:#1}}

\newcommand{\tab}[1]{Table~\ref{tab:#1}}

\newcommand{\eq}[1]{(\ref{eq:#1})}

\def\b#1{\mathbf{#1}}
\def\bb#1{\mathbb{#1}}
\def\cal#1{\mathcal{#1}}
\def\txt#1{\text{#1}}
\usepackage{xcolor}

\begin{document}

\title{\Large \bf Self-supervised Deformation Modeling for Facial Expression Editing}

\author{\parbox{16cm}{\centering
    {\large ShahRukh Athar, Zhixin Shu, and Dimitris Samaras}\\
    {\large
    Stony Brook University\\
    New York, 11794, U.S.A.}\\
    {\fontfamily{fi4}\normalsize
    \{sathar,zhshu,samaras\}@cs.stonybrook.edu}}%
}
\date{}
\thispagestyle{empty}
\pagestyle{plain}
\maketitle

\begin{abstract}

Recent advances in deep generative models have demonstrated impressive results in photo-realistic facial image synthesis and editing.  Facial expressions are inherently the result of muscle movement. However, existing neural network-based approaches usually only rely on texture generation to edit expressions and largely neglect the motion information. In this work, we propose a novel end-to-end network that disentangles the task of facial editing into two steps: a ``motion-editing'' step and a ``texture-editing'' step. In the ``motion-editing'' step, we explicitly model facial movement through image deformation, warping the image into the desired expression. In the ``texture-editing'' step, we generate necessary textures, such as teeth and shading effects, for a photo-realistic result. Our physically-based task-disentanglement system design allows each step to learn a focused task, removing the need of generating texture to hallucinate motion. Our system is trained in a self-supervised manner, requiring no ground truth deformation annotation. Using Action Units \cite{FACS} as the representation for facial expression, our method improves the state-of-the-art facial expression editing performance in both qualitative and quantitative evaluations.
\end{abstract}

\section{Introduction}

\begin{figure}[h!]
    \centering
    \renewcommand{\tabcolsep}{0.5pt}
    \begin{subfigure}{.9\linewidth}
        \centering
        \centering
        \begin{tabular}{lccccc}
        & \(\alpha=0.0\) & \(\alpha=0.2\) & \(\alpha=0.5\) & \(\alpha=0.7\) & \(\alpha=1.0\)\\
        \rotatebox{90}{\tiny{Deformed Image}} &
        \includegraphics[width=0.2\linewidth]{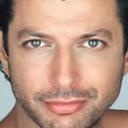}&
        \includegraphics[width=0.2\linewidth]{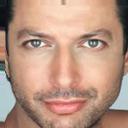}&
        \includegraphics[width=0.2\linewidth]{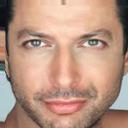}&
        \includegraphics[width=0.2\linewidth]{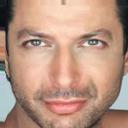}&
        \includegraphics[width=0.2\linewidth]{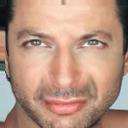}\\[-3pt]
        
        \rotatebox{90}{\hspace{7pt}\tiny{Output}} &
        \includegraphics[width=0.2\linewidth]{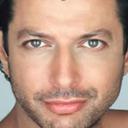}&
        \includegraphics[width=0.2\linewidth]{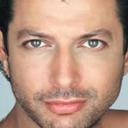}&
        \includegraphics[width=0.2\linewidth]{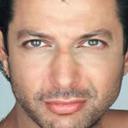}&
        \includegraphics[width=0.2\linewidth]{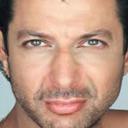}&
        \includegraphics[width=0.2\linewidth]{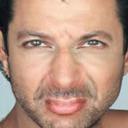}\\[-3pt]
        \end{tabular}
        \caption{Expression Editing using DefGAN}
        \label{fig:disgust_expDefGAN}
        \vspace{0.5cm}
    \end{subfigure}
    \begin{subfigure}{.9\linewidth}
        \centering
        \centering
        \begin{tabular}{lccccc}
        & \(\alpha=0.0\) & \(\alpha=0.2\) & \(\alpha=0.5\) & \(\alpha=0.7\) & \(\alpha=1.0\)\\
        \rotatebox{90}{\hspace{7pt}\tiny{Output}} &
        \includegraphics[width=0.2\linewidth]{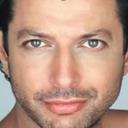}&
        \includegraphics[width=0.2\linewidth]{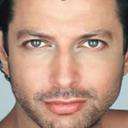}&
        \includegraphics[width=0.2\linewidth]{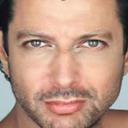}&
        \includegraphics[width=0.2\linewidth]{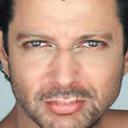}&
        \includegraphics[width=0.2\linewidth]{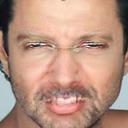}\\[-3pt]
        \end{tabular}
        \caption{Expression Editing using GANimation \cite{pumarola2018ganimation}}
        \label{fig:disgust_expGAN}
    \end{subfigure}
    \caption{
    \small{Expression Editing using DefGAN. The image on the top shows  an input face image being edited to the target expression `disgust', where \(\alpha\) controls the activation of the target AUs \cite{FACS}. DefGAN edits the image in two phases. First, in the `Motion editing' phase, we deform the input image to conform to the target expression as can be seen in the top row of \fig{disgust_expDefGAN}. Next, in the `Texture editing' phase we hallucinate the necessary textures to give us the final image (bottom row \fig{disgust_expDefGAN}). As can be seen, the deformation models facial movements and performs most of the editing work. In contrast, as shown in \fig{disgust_expGAN}, GANimation \cite{pumarola2018ganimation} edits the image entirely using texture hallucinations leading to artifacts in the final image (bottom right image of \fig{disgust_expGAN})}}    
    
    \label{fig:disgust_exp}
\end{figure}

Editing facial expressions of faces found `in-the-wild' is a problem of great interest within the Computer Vision and the Computer Graphics communities with a wide variety of applications in industries ranging from cinema to photography to e-commerce. An ideal facial expression editing system would allow its user to seamlessly change the expression of a given input face without affecting invariant attributes such as facial identity, age, etc. Recent advances in computer vision, driven by improvements in the adversarial learning framework \cite{goodfellow2014generative} have now made it possible to successfully perform such expression edits in many cases \cite{pumarola2018ganimation, StarGAN2018}.

It is  well known that facial expressions result from complex and constrained movements of facial muscles in three dimensions thus making the task of modelling and encoding them in two dimensional pixel space rather challenging. In 1978, Eckman and Friesen \cite{FACS} developed the Facial Action Coding System (FACS) that `can be used to describe any facial movement (observed in photographs, motion picture film or video tape) in terms of anatomically based action units'. Each Action Unit (AU) corresponds to a change in some specific region of the face and any anatomically possible expression can be represented as a vector of AU intensities and/or detections. For example, a smile (corresponding to happiness) can be encoded by the intensity of AU6 and AU12 and sadness can be encoded by the intensity of AU1, AU4 and AU15. Due to their interpretability and universality AUs are an ideal representation for editing expressions. 
Ground truth AU annotation of images is generally done by trained experts and is rather time consuming. Over the years however, a number of learning-based methods have been developed that are able to predict the AU activations of a face in any given image with low error rates, thus making the AU annotation of large scale datasets feasible. Recent work, such as GANimation \cite{pumarola2018ganimation} relies on AUs as a weak supervision signal, to learn a facial expression model that allows one to seamlessly transition between expressions by interpolating in the AU space. Despite its use of AUs as a supervision signal, GANimation \cite{pumarola2018ganimation} does not explicitly model facial movement and instead relies on texture synthesis to mimic facial movement effects. A downside of this is that there can be significant artifacts in the editing results such as the disappearing eyebrows and changes in the beard  that can be seen in \fig{disgust_exp}. %

In this work, we propose DefGAN, a method that separates the task of editing expressions into two sequential phases - a `Motion Editing' phase which models facial movements as an image deformation followed by a `Texture Editing' phase to hallucinate final details that arise from the appearance or disappearance of texture (such as teeth). In the `Motion Editing' phase, a Convolutional Neural Network (CNN) models facial movement explicitly by predicting a deformation field that deforms the input face to better conform to the target expression (for example curving the region around the mouth into a crescent to generate a smile). We leverage recent work from \cite{shu2018deforming} and model facial movements using an offset based deformation field. In the `Texture Editing' phase, we hallucinate the necessary textures (such as teeth, shadows) using another CNN on top of the deformed image which completes the editing process and gives us the final edited image. 

We weakly supervise our networks using  available, easily obtailnable,  AU annotations on the EmotionNet dataset \cite{emotionet} and use an adversarial learning framework similar to \cite{pumarola2018ganimation,StarGAN2018} for training. The generator is tasked with editing a given input image towards a desired target expression while ensuring the edited image can be mapped back to the original image using a cyclic transformation \cite{pumarola2018ganimation, StarGAN2018, CycleGAN2017}, while the discriminator ensures that the edited image looks real and also conforms to the target expression. 

To summarize, we develop a method that learns to edit expressions by learning anatomically consistent expression-conditioned deformation fields (without ground-truth deformation annotations) through the explicit disentanglement of the editing process into `Motion editing' and `Texture editing' phases.
The user studies we have conducted show that facial expression editing methods such as GANimation \cite{pumarola2018ganimation} and ours  produce realistic expressions (with an average plausible score of 3.5 out of a maximum score of 4.0) with the scores having a bimodal distribution. Encouraged by these results, we carried out a user study to directly compare the quality of DefGAN's editing results with GANimation \cite{pumarola2018ganimation} and found that, on average, users prefer the editing results of DefGAN over GANimation \cite{pumarola2018ganimation}. In addition to the user study, we also carry extensive expressions edits on a large variety of faces in-the-wild and show that editing expressions by explicit disentanglement of facial movement and texture synthesis leads to more realistic results and while better preserving expression-invariant features of the face. 

\section{Related Work}
Over the past few years there has been a significant amount of work on editing expression and more broadly in transferring images from one domain to another. In this section we discuss work that is most relevant to ours.
\paragraph{Face Manipulation.} Extensive work has been done on face manipulation and expression editing within the field computer vision, computer graphics and machine learning. The earliest face expression models relied on mass-and-skin models to model facial movement \cite{fischler1973representation}, such models however could not model finer skin movements that are often involved in facial expressions. Another line of research used registered 3D scans of faces to linear low-dimensional embedding of the face \cite{blanz1999morphable} by explicitly taking into account variations due to expressions. Though such models often produce realistic expressions edits, they require expensive 3D scans of the same person with different expressions and cannot be easily scaled to learn from larger datasets. Other work using detailed 3D Scans of the human face to edit expressions include 
\cite{WangCGF2004,Vlasic:2005:FTM,Song:4351915}.

Recent developments in generative adversarial networks have allowed the training of Convolutional Neural Networks to not only generate face images with great photorealism \cite{StyleGAN} but have also led to the development of a number of unsupervised and weakly supervised facial manipulation methods such as \cite{GAGAN, NeuralFace2017}. More specifically, in the case of expression editing, adversarial learning has made possible the use of landmarks \cite{GeoGANexp}, discreet expression labels \cite{StarGAN2018} and continuous Action Units \cite{pumarola2018ganimation} as a source of weak supervision. In this work, we choose to use Action Units \cite{FACS} as a source of weak supervision due their interpretability (each AU corresponds to a change in some region of the face) and wide applicability (AUs can encode any anatomically possible facial expression \cite{FACS}).
\paragraph{Generative Adversarial Networks.} Generative Adversarial Networks (GANs) \cite{goodfellow2014generative} are a powerful class of generative models that have essentially become standard within the computer vision community for unconditional image generation. GANs working by pitting neural networks against each other; the generator network is tasked with producing samples that are indistinguishable from the real data distribution while the discriminator is tasked with distinguishing between the samples generated by the generator and the real data. Follow up work \cite{WGAN, gulrajani2017improved}, have significantly improved the training stability of GANs by minimizing the Wasserstien Distance instead of the Jenson-Shanon divergence between distributions. Ever since their inception, GANs have used for a wide variety of computer vision tasks ranging from image inpainting \cite{IizukaSIGGRAPH2017, GAN:inpainting} to super-resolution \cite{SRGAN}.
\paragraph{Conditonal GANs.} Conditional GANs are a subset of GANs that use some model conditonal distributions instead on unconditonal distributions. Conditional GANs have been incredibly successfull in a variety of computer vision tasks such as image in-painting \cite{IizukaSIGGRAPH2017}, super-resolution \cite{SRGAN}, and domain transfer \cite{CycleGAN2017, StarGAN2018,kim2017learning}.
\paragraph{Deformation modelling.} Over the past few years there has been a growing interest in learning deformations from large scale datasets. Work done in \cite{shu2018deforming} shows that it is possible to model face images as deformations of texture templates in a completely unsupervised manner.  Early work on using deformations to edit expressions includes Expression flow \cite{Yang:2011:EFF:1964921.1964955} which edits expressions by warping the input image using a 2D flow field. This flow field is estimated using another image of the same person with the desired expression. This work builds on prior work on incorporating deformations within convolutional networks such as \cite{jaderberg2015spatial,dai2017deformable}.
\paragraph{Unpaired Image-to-Image Translation.} We cast the problem of editing expressions in the framework of unpaired image to image translation where the target image is the image of the person in the input with the target expression. The advent of GANs \cite{goodfellow2014generative} have made it possible to produce incredibly photorealistic results from when translating from one domain to another. For example, pix2pix \cite{pix2pix2016} is able to generate high quality images object from mere sketches or segmentation maps as inputs. More relevant to this paper is work like \cite{CycleGAN2017,StarGAN2018,kim2017learning,pumarola2018ganimation} which can be used to transfer between  various attributes of the face including including expressions.
\section{Method}

\begin{figure*}[h]
    \centering
    \includegraphics[width=\textwidth]{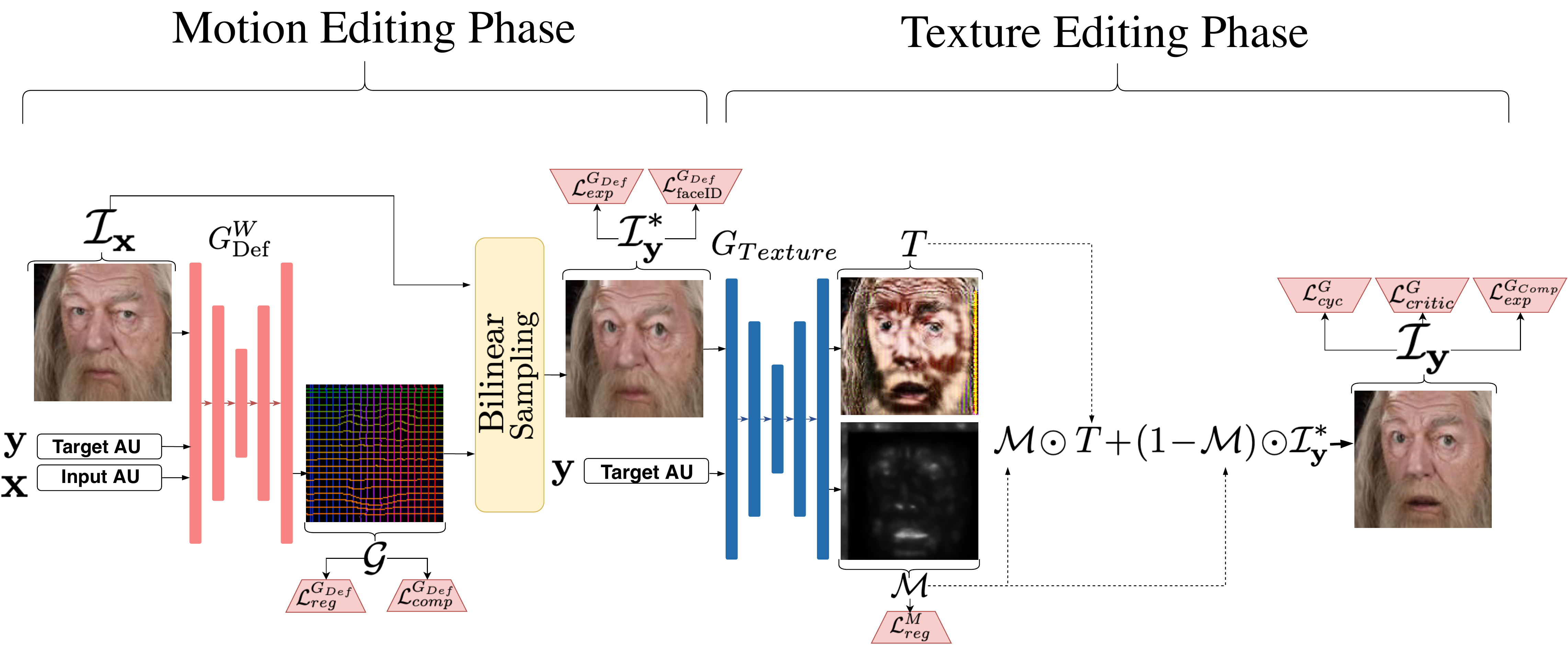}
    \caption{\small{The Architecture of DefGAN and its associated losses. DefGAN edits expressions in two phases. First, in the `Motion editing' phase, DefGAN models facial movement by predicting a deformation field and deforms the input image using it (note the widening of the eyes in \(\mathcal{I}_{\mathbf{y}}^{*}\)). Next, in the `Texture Editing' phase we hallucinate the necessary textures to complete the editing process (note the opening of the mouth in \(\mathcal{I}_{\mathbf{y}}\)).}}
    \label{fig:system}
\end{figure*}

Consider an input image \(\cal{I}_{\b{x}} \in \bb{R}^{H \times W \times 3}\) with some expression \(\b{x}\). We'd like to change the expression of the person in \(\cal{I}_{\b{x}}\) to some target expression \(\b{y}\) to give us \(\cal{I}_{\b{y}}\). Here, the expressions are encoded by AU \cite{FACS} intensities, \(\b{x} = (x_{1},\hdots,x_{n})\), where each \(x_{i}\) is the intensity of the \(i^{\text{th}}\) AU scaled between 0 and 1. To carry out this expression edit we transform the image in two stages. First, in the motion editing phase, we deform the input image - modeling facial movement in the pixel space - to conform to the target expression. More specifically, the first stage can be written as follows: 
\vspace{-1pt}
\begin{equation}
    \begin{split}
        \cal{I}_{\b{y}}^{*} = G_{\text{Def}}^{W}(\cal{I}_{\b{x}}, \b{x}, \b{y})
    \end{split}
    \label{eq:def}
\end{equation}

\noindent where, \(G_{\text{Def}}^{W}\) is the deformation generator that takes as input \(\cal{I}_{\b{x}}\), the input expression \(\b{x}\) and target expression \(\b{y}\) and produces the deformed image \(\cal{I}_{\b{y}}^{*}\). Although this image could have been appropriately deformed to achieve the target expression \(y\), it might still lack the necessary texture modifications to look realistic. For example, if we had to edit a face with a neutral expression to a face with a grin, the best a deformed image could give us is a very wide (and possibly unrealistic) smile, we'd still need to hallucinate the texture of the teeth to get the correct target expression. We hallucinate the necessary texture, in the texture editing phase, using another convolutional network as follows:
\vspace{-1pt}
\begin{equation}
     \cal{I}_{\b{y}} = G_{\text{Texture}}(\cal{I}_{\b{y}}^{*},  \b{y})
     \label{eq:hal}
\end{equation}
\noindent where, \(G_{\text{Texture}}\) is the texture hallucination network. In the interest of brevity, we denote the entire transformation from input image to the final image i.e \(\cal{I}_{\b{x}} \rightarrow \cal{I}_{\b{y}}\), as follows:
\begin{equation}
    \cal{I}_{\b{y}} = G_{\text{Comp}}(\cal{I}_{\b{x}}, \b{x}, \b{y})
    \label{eq:comp}
\end{equation}
\noindent where, \(G_{\text{Comp}} \coloneqq G_{\text{Texture}} \odot  G_{\text{Def}}^{W}\) denotes the composition of \eq{def} and \eq{hal}.

\subsection{Architecture}
DefGAN consists of three convolutional networks. The deformation generator \(G_{\text{Def}}^{W}\), the texture hallucination network \(G_{\text{Texture}}\), and a discriminator \(D\) with a critic output, \(D_{c}\), and an AU regression output, \(D_{exp}\).
\subsubsection{Motion Editing Phase}
The first stage of DefGAN's expression editing involves deforming the input image, \(\cal{I}_{\b{x}}\) to \(\cal{I}_{\b{y}}^{*}\) such that \(\cal{I}_{\b{y}}^{*}\) closely approximates the target expression. We carry out this transformation using a deformation generator \(G_{\text{Def}}^{W}\) that first predicts a deformation grid and then warps the input image using it. More specifically, the transformation can be written as follows
\begin{equation}
    \begin{split}
        \cal{G} &= G_{\text{Def}}(\cal{I}_{\b{x}}, \b{x}, \b{y})\\
        \cal{I}_{\b{y}}^{*} &= \text{Warp}(\cal{G},\cal{I}_{\b{x}})
    \end{split}
    \label{eq:def_split}
\end{equation}
\noindent where, \(\cal{G}\) is the predicted deformation grid. The deformation generator, \(G_{\text{Def}}^{W}\), is just the composition of the above two operations
\vspace{-1pt}
\begin{equation}
\begin{split}
    G_{\text{Def}}^{W}(\cal{I}_{\b{x}}, \b{x}, \b{y}) &\coloneqq \text{Warp}(G_{\text{Def}}(\cal{I}_{\b{x}}, \b{x}, \b{y}),\cal{I}_{\b{x}})\\
    \cal{I}_{\b{y}}^{*} &= G_{\text{Def}}^{W}(\cal{I}_{\b{x}}, \b{x}, \b{y})
\end{split}
\end{equation}
We use an offset based deformation grid as proposed in \cite{shu2018deforming} with a maximum offset of 5 pixels.

\subsubsection{Texture Editing Phase}
The second stage of DefGAN's expression edit, the texture editing phase, involves hallucinating the necessary features that cannot be modelled by a deformation to improve the realism of the final image and its fidelity to the target expression. Like prior work \cite{pumarola2018ganimation}, we found that a masking mechanism helps texture generator create better quality edits
\begin{equation}
\begin{split}
    \cal{T}, \cal{M} &= G_{\text{Texture}}^{*}(\cal{I}_{\b{y}}^{*},  \b{y}) \\
    \cal{I}_{\b{y}} &= \cal{M} \odot T + (1 - \cal{M}) \odot  \cal{I}_{\b{y}}^{*}
\end{split}
\end{equation}
\noindent where, \(\cal{T}\) is the hallucinated texture map and \(\cal{M}\) is the attention mask. The texture network, \(G_{\text{Texture}}\) is just the composition of the above two operations.

\subsection{Training}
 Similar to prior work \cite{pumarola2018ganimation, StarGAN2018} we rely on a GAN  based framework \cite{goodfellow2014generative} to train the deforming network, \(G_{\text{Def}}^{W}\) and the texture network, \(G_{\text{Texture}}\) jointly. In addition to an adversarial loss, we train DefGAN to also minimize an expression loss, a cycle consistency loss, a facial identity loss and a regularization loss on the deformation grid.
 
 \paragraph{Adversarial Loss.} In order to ensure DefGAN's image edit looks natural we train the generators to minimize an adversarial loss \cite{goodfellow2014generative}. Instead of using the standard GAN loss, which corresponds to minimizing the Jenson-Shannon divergence, we use the WGAN-GP loss \cite{gulrajani2017improved} which minimizes the Earth Mover Distance between the generated and the real distribution. Specifically, let  \(\cal{I}_{\b{x}}\) be the input image with expression \(\b{x}\), let \(\b{y}\) be the target expression and let \(\cal{P}_{r}\) be the real distribution of images. The critic loss for the discriminator, \(D\) is given as follows
 \vspace{-1pt}
 \begin{equation}
     \begin{split}
         \cal{L}_{critic}^{D} = &\bb{E}_{\cal{I}_{\b{x}} \sim \cal{P}_{r}}\left[ D_{c}\left(G_{\text{Comp}}(\cal{I}_{\b{x}}, \b{x}, \b{y})\right)\right]\\ 
         &-  \bb{E}_{\cal{I}_{\b{x}} \sim \cal{P}_{r}}\left[ D_{c}\left(\cal{I}_{\b{x}})\right)\right]\\ 
         &+ \lambda_{gp}\bb{E}_{\hat{\cal{I}} \sim \hat{\cal{P}}}\left[(\|\nabla_{\hat{\cal{I}}}D_{c}(\hat{\cal{I}})\|_{2} - 1)^{2}\right]
     \end{split}
     \label{eq:D_critic}
 \end{equation}
 
 Where, \(D_{c}\) is the critic output of the discriminator \(D\), \(\lambda_{gp}\) is the gradient penalty coefficient and \(\hat{\cal{P}}\) is the interpolated distribution. The generators, \( G_{\text{Def}}^{W} \text{ and }  G_{\text{Texture}}\) are trained to `please' the critic by maximizing the critic score. We express  this  loss as:
 \vspace{-1pt}
 \begin{equation}
     \cal{L}_{critic}^{G} = -\bb{E}_{\cal{I}_{\b{x}} \sim \cal{P}_{r}}\left[ D_{c}\left(G_{\text{Comp}}(\cal{I}_{\b{x}}, \b{x}, \b{y})\right)\right]
     \label{eq:G_critic}
 \end{equation}
 
 \paragraph{Expression Loss.} To ensure that the generators, while producing a realistic image, are also generating the correct target expression we add a loss that penalizes deviations from the target expression. This loss is defined using the AU output of the discriminator, \(D_{exp}\), that is trained to predict the AU intensities for any given input image \(\cal{I}_{\b{x}}\). The loss is defined as follows:
 \vspace{-1pt}
 
 \begin{equation}
    \begin{split}
     &\cal{L}_{exp}^{G_{Def}} = \lambda_{exp}^{G_{Def}}\underset{{\cal{I}_{\b{x}} \sim \cal{P}_{r}}}{\bb{E}}\left[\|D_{exp}(G_{\text{Def}}(\cal{I}_{\b{x}}, \b{x}, \b{y})) - \b{y}\|_{2}^{2}\right] \\
     &\cal{L}_{exp}^{G_{Comp}} = \lambda_{exp}^{G_{Comp}}\underset{\cal{I}_{\b{x}} \sim \cal{P}_{r}}{\bb{E}}\left[\|D_{exp}(G_{\text{Comp}}(\cal{I}_{\b{x}}, \b{x}, \b{y})) - \b{y}\|_{2}^{2}\right] \\
     &\cal{L}_{exp}^{G} = \cal{L}_{exp}^{G_{Def}} + \cal{L}_{exp}^{G_{Comp}}\\  
     \label{eq:G_exp}
    \end{split}
 \end{equation}
 \vspace{-1pt}
 Here, \(\lambda_{exp}^{G_{Comp}} \text{ and }\lambda_{exp}^{G_{Def}}\) are the coefficients of each term.
 We apply the expression loss both on the final image output \(\cal{I}_{\b{y}} = G_{\text{Comp}}(\cal{I}_{\b{x}}, \b{x}, \b{y})\) and the intermediate image output \(\cal{I}_{\b{y}}^{*} = G_{\text{Def}}^{W}(\cal{I}_{\b{x}}, \b{x}, \b{y})\).
 The AU output, \(D_{exp}\), is trained to minimize the AU prediction error on real images
 \vspace{-1pt}
 \begin{equation}
     \cal{L}_{exp}^{D} = \bb{E}_{\cal{I}_{\b{x}} \sim \cal{P}_{r}}\left[\|D_{exp}(\cal{I}_{\b{x}}) - \b{x}\|_{2}^{2}\right]
     \label{eq:D_exp}
 \end{equation}
 
 \paragraph{Cycle Loss.} In order to preserve subject identity as DefGAN edits the image,  we enforce a cycle consistency loss on the generator networks as follows:
 \begin{equation}
     \begin{split}
         \cal{L}_{cyc}^{G} =  \lambda_{cyc}\bb{E}_{\cal{I}_{\b{x}} \sim \cal{P}_{r}}\left[\|G_{\text{Comp}}(G_{\text{Comp}}(\cal{I}_{\b{x}}, \b{x}, \b{y}),\b{y},\b{x}) - \cal{I}_{\b{x}}\|_{1}\right]
     \end{split}
     \label{eq:G_cyc}
 \end{equation}
 
 In the absence of ground truth images for each person with different annotated expressions, we found that this cycle loss ensures the identity of the person does not change as the expression changes.
 
 \paragraph{Face Identity Loss.} We regularize the deformation generator,  \(G_{\text{Def}}^{W}\), by ensuring that it preserves the identity of the person as it deforms the input image. More specifically, we maximize the cosine similarity between the OpenFace \cite{amos2016openface} embedding of the deformed input image, \( \cal{I}_{\b{y}}^{*}\), and the input image \( \cal{I}_{\b{x}}\). This  loss can be expressed as 
 \begin{equation}
      \cal{L}_{\txt{faceID}}^{G_{Def}} =  \bb{E}_{\cal{I}_{\b{x}} \sim \cal{P}_{r}}\left[(1 - \text{cos}(\text{OpenFace}( \cal{I}_{\b{y}}^{*}), \text{OpenFace}( \cal{I}_{\b{x}}))\right]
      \label{eq:face_id}
 \end{equation}

\paragraph{Composition Loss. } We found that imposing a composition loss on the generated deformation grid \(\cal{G}\) was useful in producing realistic expression edits. The grid composition loss is defined as follows:
\begin{equation}
     \begin{split}
          \cal{L}_{comp}^{G_{Def}} &=   \lambda_{comp}\bb{E}_{\cal{I}_{\b{x}} \sim \cal{P}_{r}}\left[(\text{Warp}(\cal{G}_{cyc}, \cal{G}) - \cal{I}_{Def})^{2}\right]\\
     \end{split}
      \label{eq:reg}
\end{equation}
\noindent where, \(\cal{G}_{cyc}\) is the deformation grid produced during the cycle transformation and \(\cal{I}_{Def}\) is the identity deformation grid.

 \paragraph{Regularization.} To ensure smoothness of the generated deformation fields we add a TV-regularization term on the deformation grid, \(\cal{G}\) as defined in \eq{def_split}, and also penalize the difference between \(\cal{G}\) and the identity deformation. The regularization terms can be written as follows
 
 \begin{equation}
     \begin{split}
          &\cal{L}_{reg}^{G_{Def}} =   \lambda_{eye}^{\cal{G}}\bb{E}_{\cal{I}_{\b{x}} \sim \cal{P}_{r}}\left[(\cal{G} - \cal{I}_{Def})^{2}\right]\\
          &+ \lambda_{TV}^{G}\bb{E}_{\cal{I}_{\b{x}} \sim \cal{P}_{r}}\left[\sum_{i,j}^{H,W}(\cal{G}_{i+1,j} - \cal{G}_{i+1,j})^{2} + (\cal{G}_{i,j+1} - \cal{G}_{i,j})^{2}\right]
     \end{split}
      \label{eq:reg}
 \end{equation}
 \noindent where, \(\cal{I}_{Def}\) is the identity deformation grid.
 We also add a similar regularization to the mask, \(\cal{M}\) that is generated during the texture editing phase. This regularization term is as follows
 \begin{equation}
    \hspace{-1pt}
     \begin{split}
          &\cal{L}_{reg}^{M} =   \lambda_{eye}^{M}\bb{E}_{\cal{I}_{\b{x}} \sim \cal{P}_{r}}\left[\|\cal{M}\|_{2}^{2}\right]\\
          &+ \lambda_{TV}^{\cal{M}}\bb{E}_{\cal{I}_{\b{x}} \sim \cal{P}_{r}}\left[\sum_{i,j}^{H,W}(\cal{M}_{i+1,j} - \cal{M}_{i+1,j})^{2} + (\cal{M}_{i,j+1} - \cal{M}_{i,j})^{2}\right]
     \end{split}
      \label{eq:reg_mask}
 \end{equation}
 
 \paragraph{Final Loss.} The total loss on the generators is
 \begin{equation}
    \cal{L}_{Total}^{G} =  \cal{L}_{critic}^{G} + \cal{L}_{exp}^{G} + \cal{L}_{cyc}^{G} + \cal{L}_{\txt{faceID}}^{G_{Def}} + \cal{L}_{comp}^{G_{Def}} + \cal{L}_{reg}^{G_{Def}} + \cal{L}_{reg}^{M}
    \label{eq:loss_g}
 \end{equation}
 We minimize this loss over the parameters of \( G_{\text{Def}}\txt{ and } G_{\text{Texture}}\) to convergence
 \begin{equation}
     \begin{split}
          G_{\text{Def}}^{W*}, G_{\text{Texture}}^{*} = \underset{ G_{\text{Def}}^{W}, G_{\text{Texture}}}{\txt{argmin}} \cal{L}_{Total}^{G}
     \end{split}
 \end{equation}
 The total loss on the discriminator is 
 \begin{equation}
     \begin{split}
         \cal{L}_{Total}^{D} = \cal{L}_{critic}^{D} + \cal{L}_{exp}^{D}
     \end{split}
     \label{eq:loss_d}
 \end{equation}
 The discriminator is trained to minimize this loss
 \begin{equation}
     \begin{split}
          D^{*} = \underset{ D}{\txt{argmin }} \cal{L}_{Total}^{D}
     \end{split}
 \end{equation}

\section{Experiments and Results}

We evaluated our model on a variety of facial identities and on a range of expression editing tasks to test its quality and robustness. Around 170 different facial identities from the CelebA-HQ dataset \cite{Karras18} were used for evaluation. In addition to CelebA-HQ \cite{Karras18} we scraped 40 more images from the internet with variations in pose, illumination and facial attributes to test the robustness of our model on more challenging input images. First, we show the results of editing expressions on a number of in-the-wild faces and measure the change in facial identity after the expression edit by comparing the OpenFace \cite{amos2016openface} embeddings of the edited image and the input image. Next, we show the results of manipulating single Action Units \cite{FACS} using our model and we finally discuss the results of a user study conducted to determine which model among DefGAN and GANimation \cite{pumarola2018ganimation} produced better expression edits on in-the-wild images as judged by humans.

\subsection{Training Details}
\paragraph{DefGAN.}DefGAN was trained on a subset of the EmotionNet dataset \cite{emotionet} containing 190k images. We use the Adam optimizer \cite{Adam} with an initial learning rate of \(1e-4\), \(\beta_{1} = 0.5\), \(\beta_{2} = 0.999\) and a batch size of 25. The model was trained for 40 epochs with the learning rate decaying to 0 over the last 20 epochs. The deformation generator and the texture generator were optimized jointly. The critic was trained for 10 steps for every step of the generators.
\paragraph{GANimation.} GANimation \cite{pumarola2018ganimation} was trained on the same subset of the EmotionNet dataset \cite{emotionet} containing 190k images. We used the  hyperparameters  used by the authors in \cite{pumarola2018ganimation} with the only difference that we train for 40 epochs.

\subsection{Expression Synthesis on Wild Faces}

\begin{figure}[h!]
    \centering
    \includegraphics[width=0.8\linewidth]{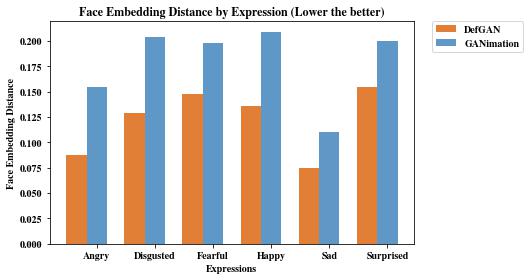}
    \caption{\small{\textbf{Face Embedding Distance.}  Here we show the distance between the CMU-OpenFace \cite{amos2016openface} embeddings of the input image and the images edited by DefGAN and GANimation. As one can see, DefGAN's edits consistently preserve facial identity better than GANimations's. 
    }}
    \label{fig:FaceID_Loss}
\end{figure}

\begin{figure*}[h!]
    \centering
    \renewcommand{\tabcolsep}{0.5pt}
    \begin{tabular}{lccccccc}
     & \small{Input} & \small{Anger} & \small{Disgust} & \small{Fear} & \small{Happy} & \small{Sad} & \small{Surprise}\\
    \rotatebox{90}{\hspace{15pt}\scriptsize{DefGAN}} &
    \includegraphics[width=0.12\textwidth]{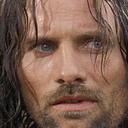}&
    \includegraphics[width=0.12\textwidth]{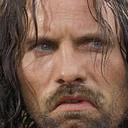}&
    \includegraphics[width=0.12\textwidth]{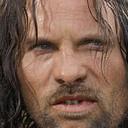}&
    \includegraphics[width=0.12\textwidth]{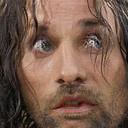}&
    \includegraphics[width=0.12\textwidth]{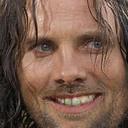}&
    \includegraphics[width=0.12\textwidth]{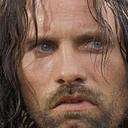}&
    \includegraphics[width=0.12\textwidth]{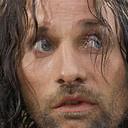}\\[-2pt]
    \rotatebox{90}{\scriptsize{\hspace{0pt}GANimation \cite{pumarola2018ganimation}}} &
    \includegraphics[width=0.12\textwidth]{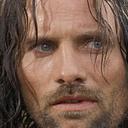}&
    \includegraphics[width=0.12\textwidth]{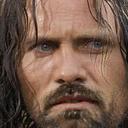}&
    \includegraphics[width=0.12\textwidth]{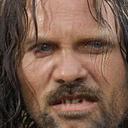}&
    \includegraphics[width=0.12\textwidth]{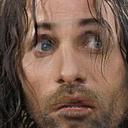}&
    \includegraphics[width=0.12\textwidth]{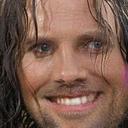}&
    \includegraphics[width=0.12\textwidth]{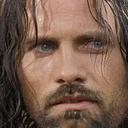}&
    \includegraphics[width=0.12\textwidth]{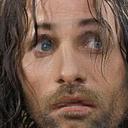}\\[0pt]
    \rotatebox{90}{\hspace{15pt}\scriptsize{DefGAN}} &
    \includegraphics[width=0.12\textwidth]{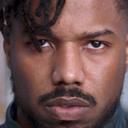}&
    \includegraphics[width=0.12\textwidth]{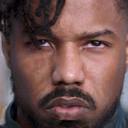}&
    \includegraphics[width=0.12\textwidth]{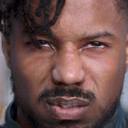}&
    \includegraphics[width=0.12\textwidth]{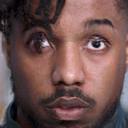}&
    \includegraphics[width=0.12\textwidth]{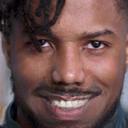}&
    \includegraphics[width=0.12\textwidth]{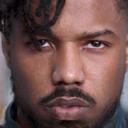}&
    \includegraphics[width=0.12\textwidth]{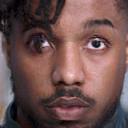}\\[-2pt]
    \rotatebox{90}{\scriptsize{\hspace{0pt}GANimation \cite{pumarola2018ganimation}}} &
    \includegraphics[width=0.12\textwidth]{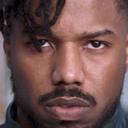}&
    \includegraphics[width=0.12\textwidth]{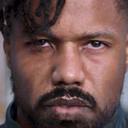}&
    \includegraphics[width=0.12\textwidth]{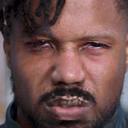}&
    \includegraphics[width=0.12\textwidth]{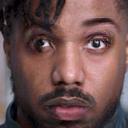}&
    \includegraphics[width=0.12\textwidth]{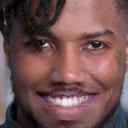}&
    \includegraphics[width=0.12\textwidth]{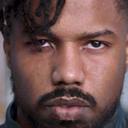}&
    \includegraphics[width=0.12\textwidth]{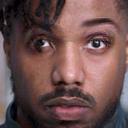}\\[1pt]
    \rotatebox{90}{\hspace{15pt}\scriptsize{DefGAN}} &
    \includegraphics[width=0.12\textwidth]{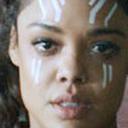}&
    \includegraphics[width=0.12\textwidth]{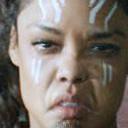}&
    \includegraphics[width=0.12\textwidth]{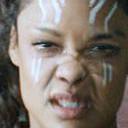}&
    \includegraphics[width=0.12\textwidth]{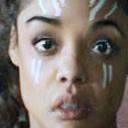}&
    \includegraphics[width=0.12\textwidth]{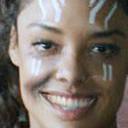}&
    \includegraphics[width=0.12\textwidth]{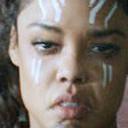}&
    \includegraphics[width=0.12\textwidth]{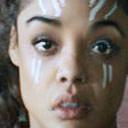}\\[-2pt]
    \rotatebox{90}{\scriptsize{\hspace{0pt}GANimation \cite{pumarola2018ganimation}}} &
    \includegraphics[width=0.12\textwidth]{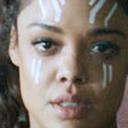}&
    \includegraphics[width=0.12\textwidth]{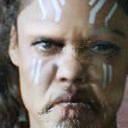}&
    \includegraphics[width=0.12\textwidth]{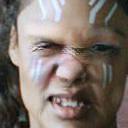}&
    \includegraphics[width=0.12\textwidth]{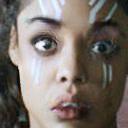}&
    \includegraphics[width=0.12\textwidth]{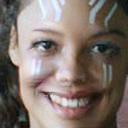}&
    \includegraphics[width=0.12\textwidth]{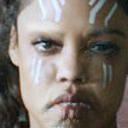}&
    \includegraphics[width=0.12\textwidth]{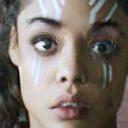}\\[0pt]

    \rotatebox{90}{\hspace{15pt}\scriptsize{DefGAN}} &
    \includegraphics[width=0.12\textwidth]{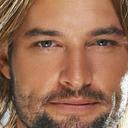}&
    \includegraphics[width=0.12\textwidth]{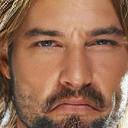}&
    \includegraphics[width=0.12\textwidth]{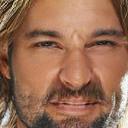}&
    \includegraphics[width=0.12\textwidth]{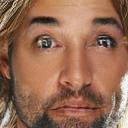}&
    \includegraphics[width=0.12\textwidth]{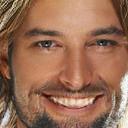}&
    \includegraphics[width=0.12\textwidth]{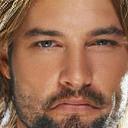}&
    \includegraphics[width=0.12\textwidth]{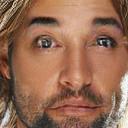}\\[-2pt]
    \rotatebox{90}{\scriptsize{\hspace{0pt}GANimation \cite{pumarola2018ganimation}}} &
    \includegraphics[width=0.12\textwidth]{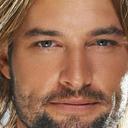}&
    \includegraphics[width=0.12\textwidth]{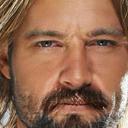}&
    \includegraphics[width=0.12\textwidth]{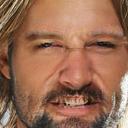}&
    \includegraphics[width=0.12\textwidth]{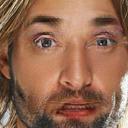}&
    \includegraphics[width=0.12\textwidth]{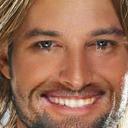}&
    \includegraphics[width=0.12\textwidth]{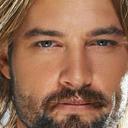}&
    \includegraphics[width=0.12\textwidth]{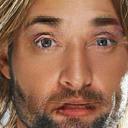}\\[-2pt]
    \end{tabular}
    \caption{\small{Editing Facial Expressions. Here we show images edited by DefGAN and GANimation \cite{pumarola2018ganimation} to various target expressions. GANimation \cite{pumarola2018ganimation} tends to produce artifacts (results of `Anger' and `Disgust' in row 3) or ends up hallucinating inaccurate textures (results of `Happy' in row 2 and results of `Anger' in row 4). In contrast, the editing results of DefGAN are more consistent with fewer artifacts and more accurate textures.\\~\\}}
    \label{fig:InTheWild}
\end{figure*}

\begin{figure}[ht]
    \centering
    \renewcommand{\tabcolsep}{0.5pt}
    \begin{tabular}{lccccc}
     & \small{Input} & \small{Disgust} & \small{Fear} & \small{Happy} & \small{Sad}\\
    \rotatebox{90}{\hspace{4pt}\tiny{DefGAN}} &
    \includegraphics[width=0.15\linewidth]{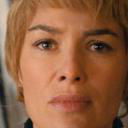}&
    \includegraphics[width=0.15\linewidth]{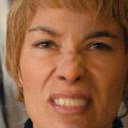}&
    \includegraphics[width=0.15\linewidth]{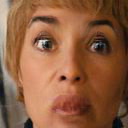}&
    \includegraphics[width=0.15\linewidth]{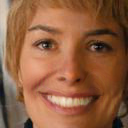}&
    \includegraphics[width=0.15\linewidth]{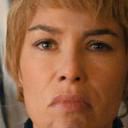}\\[-2pt]
    \rotatebox{90}{\hspace{4pt}\tiny{DefGAN}} &
    \includegraphics[width=0.15\linewidth]{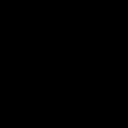}&
    \includegraphics[width=0.15\linewidth]{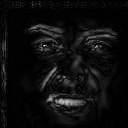}&
    \includegraphics[width=0.15\linewidth]{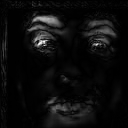}&
    \includegraphics[width=0.15\linewidth]{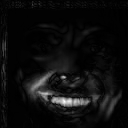}&
    \includegraphics[width=0.15\linewidth]{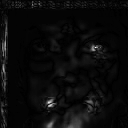}\\[-2pt]
    \rotatebox{90}{\tiny{\hspace{12pt}\cite{pumarola2018ganimation}}} &
    \includegraphics[width=0.15\linewidth]{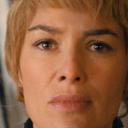}&
    \includegraphics[width=0.15\linewidth]{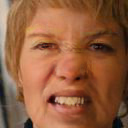}&
    \includegraphics[width=0.15\linewidth]{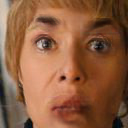}&
    \includegraphics[width=0.15\linewidth]{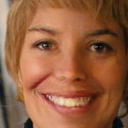}&
    \includegraphics[width=0.15\linewidth]{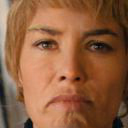}\\[-2pt]
    \rotatebox{90}{\tiny{\hspace{12pt}\cite{pumarola2018ganimation}}} &
    \includegraphics[width=0.15\linewidth]{figures/DiffImgs/zero.png}&
    \includegraphics[width=0.15\linewidth]{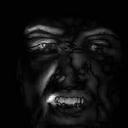}&
    \includegraphics[width=0.15\linewidth]{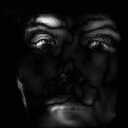}&
    \includegraphics[width=0.15\linewidth]{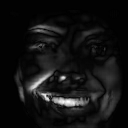}&
    \includegraphics[width=0.15\linewidth]{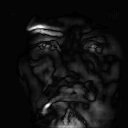}\\[0pt]
    
    \rotatebox{90}{\hspace{4pt}\tiny{DefGAN}} &
    \includegraphics[width=0.15\linewidth]{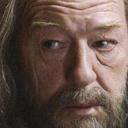}&
    \includegraphics[width=0.15\linewidth]{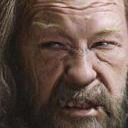}&
    \includegraphics[width=0.15\linewidth]{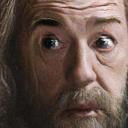}&
    \includegraphics[width=0.15\linewidth]{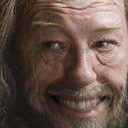}&
    \includegraphics[width=0.15\linewidth]{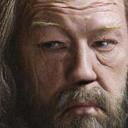}\\[-2pt]
    \rotatebox{90}{\hspace{4pt}\tiny{DefGAN}} &
    \includegraphics[width=0.15\linewidth]{figures/DiffImgs/zero.png}&
    \includegraphics[width=0.15\linewidth]{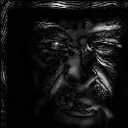}&
    \includegraphics[width=0.15\linewidth]{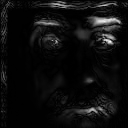}&
    \includegraphics[width=0.15\linewidth]{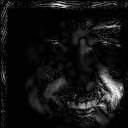}&
    \includegraphics[width=0.15\linewidth]{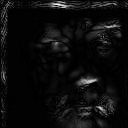}\\[-2pt]
    \rotatebox{90}{\tiny{\hspace{10pt}\cite{pumarola2018ganimation}}} &
    \includegraphics[width=0.15\linewidth]{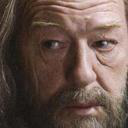}&
    \includegraphics[width=0.15\linewidth]{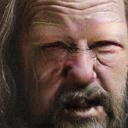}&
    \includegraphics[width=0.15\linewidth]{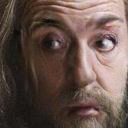}&
    \includegraphics[width=0.15\linewidth]{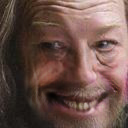}&
    \includegraphics[width=0.15\linewidth]{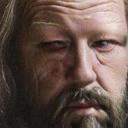}\\[-2pt]
    \rotatebox{90}{\tiny{\hspace{12pt}\cite{pumarola2018ganimation}}} &
    \includegraphics[width=0.15\linewidth]{figures/DiffImgs/zero.png}&
    \includegraphics[width=0.15\linewidth]{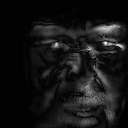}&
    \includegraphics[width=0.15\linewidth]{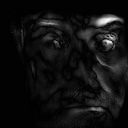}&
    \includegraphics[width=0.15\linewidth]{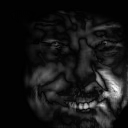}&
    \includegraphics[width=0.15\linewidth]{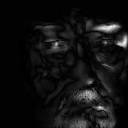}\\[-2pt]
     \end{tabular}
    \caption{\small{Difference Images. This figure shows the pixel-wise absolute difference between images edited by GANimation and the input image. As one can see, DefGAN only changes the parts of the input that are required to attain the target expression while GANimation \cite{pumarola2018ganimation} changes larger portions of the input image regardless of target expression.}}
    \label{fig:Diff}
\end{figure}

\begin{table}[h]
\centering
\begin{tabular}{@{}ll@{}}
\toprule
Method Name     & FID Score  \\ \midrule
GANimation [21] & 4.75 \\
DefGAN     & \textbf{3.82} \\ \bottomrule \\[-2pt]
\end{tabular}
\caption{\small{FID Scores of GANimation \cite{pumarola2018ganimation} and DefGAN.}}
\label{tab:FID}
\end{table}
We first tested our method by performing expression edits on in-the-wild images from CelebA-HQ dataset \cite{Karras18} and 40 more images scraped from the internet. The AU representation for each expression was computed by running the OpenFace AU detector \cite{baltrusaitis2018openface} on all the peak expression images of a randomly selected person from the MUG dataset \cite{MUG}; which consists of seven labelled expressions (anger, disgust, fear, neutral, happy, sad and surprise) for 84 persons. \fig{InTheWild} shows the results of expression edits performed on a few images from the internet and from CelebA-HQ \cite{Karras18}. As can be seen, our model consistently performs better edits across all expressions and faces. In particular, we noticed that GANimation \cite{pumarola2018ganimation} tends to distort the face by either producing artifacts (for example, the result of `Disgust' in row 3 of \fig{InTheWild}) or by `over-editing' (for example, the results of `Happy' in row 1, 2 and 4 of \fig{InTheWild}) which we posit is due to its complete reliance on the hallucination mechanism. In contrast, DefGAN, due its use of a deformation to warp the face to conform to the target expression, only hallucinates the necessary details and does not produce such artifacts. The Fr\'echet Inception Distance (FID) Score \cite{FIDNeurips2017} has become a standard measure to evaluate the realism of generative models. Lower the FID score of a model the more realistic are its images. \tab{FID} shows the FID \cite{FIDNeurips2017} score of the images edited by DefGAN and GANimation, and as one can see this further suggests that the editing results of DefGAN are more realistic than those of GANimation.

\fig{Diff} shows the absolute pixel-wise difference between the input image and the edited image across a few target expressions. Ideally, we'd only see differences in regions that correspond to the expression change. As can be seen in \fig{Diff}, the edits made by DefGAN are significantly more concentrated to the regions relevant to the final expression than the edits made by GANimation  which tend to be more spread out. \fig{FaceID_Loss} shows the distance between the edited image and the input image in the OpenFace \cite{amos2016openface} embedding space on fifty different randomly chosen representations of each expression.  DefGAN retains the input facial identity  much better than GANimation across all the six target expressions. The retention of facial identity can also be seen visually in \fig{InTheWild},  where attributes such as the beard and eyebrows of the edited images (results of `Disgust' in all rows, the results of `Happy' in row 2) have greater fidelity to the input with DefGAN's edits than GANimation's  edits which tend to either erase  or thicken them. 
\fig{disgust_exp} shows how deformations can be especially helpful in certain expression edits, such as going from a close to neutral expression to a `disgust' expression. In this expression transformation we can see that the deformation in DefGAN does most of the work of converting the input face to the `disgust' expression while the hallucination only adds minor details to the final image.

\subsection{Learnt facial movements conditioned on Action Units}

\begin{figure}[h]
    \centering
    \renewcommand{\tabcolsep}{0.5pt}
    \begin{tabular}{cc}
    \includegraphics[width=0.495\linewidth]{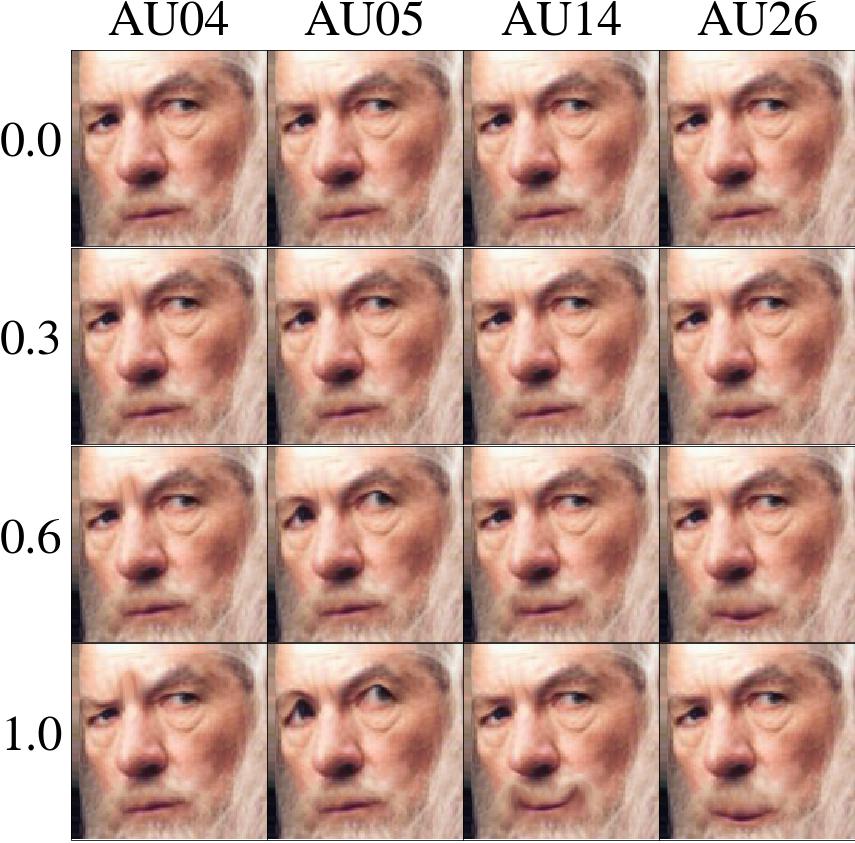}&
    \includegraphics[width=0.458\linewidth]{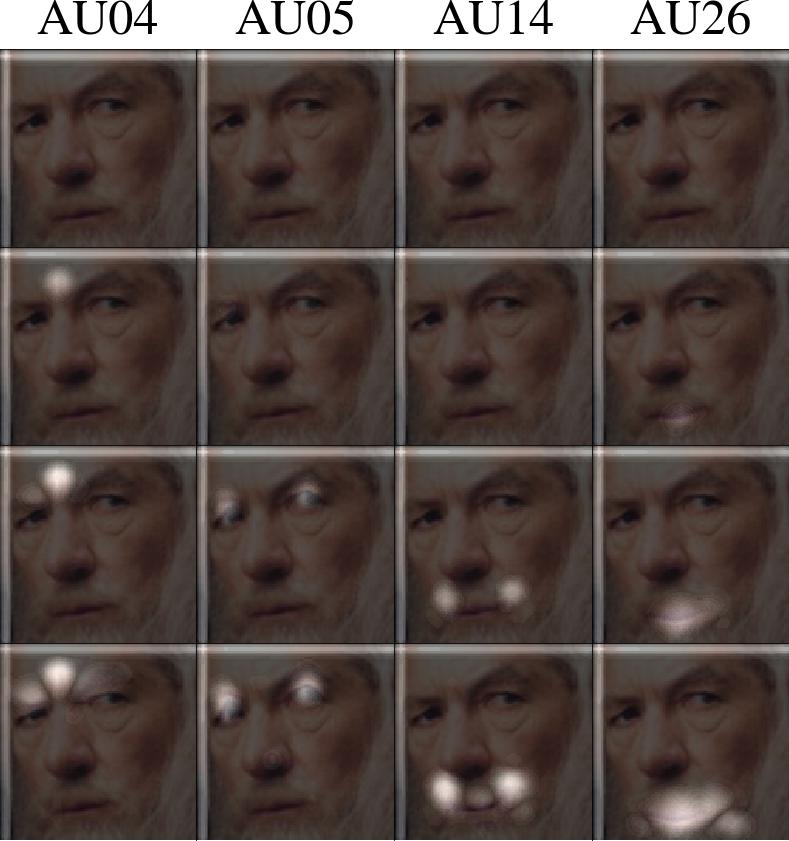}
    \end{tabular}
    \caption{\small{Learnt Action Unit conditioned facial movements. \textbf{Left:} Here we show the effects of single AU activations on the deformed image \(\cal{I}_{\b{y}}^{*}\). \textbf{Right:} Here we show the regions of the face affected by the change in intensity of the corresponding AU. As can be seen, changing the intensity of any particular AU causes smooth changes in the corresponding facial regions akin to the results of true facial muscle movement as encoded by that AU. }}
    \label{fig:AUIntDef}
\end{figure}

\vspace{-1pt}
In this section, we analyze the effect of changing individual AUs (AU14, AU5, AU14, and AU26) on an input face. We show the effect both on the deformed image. As can be seen from \fig{AUIntDef}, DefGAN successfully learns to faithfully model facial movement through its deformation mechanism. For example, when increasing the intensity of AU5 (Upper Lid Raiser) we clearly see the eyebrows raising up while other regions of the face remain unchanged. Coherent facial movements resulting from deformations can also be seen as we change the intesities of AU4 (Brow Lowerer), AU14 (Dimpler) and AU26 (Jaw Drop). Examples of more AU activations along with results of the final edited images  can be found in the appendix.

\subsection{User Study}
\label{subsec:userstudy}
\begin{figure}[h]
    \centering
    \renewcommand{\tabcolsep}{0.5pt}
    \begin{tabular}{c}
    \includegraphics[width=0.5\linewidth]{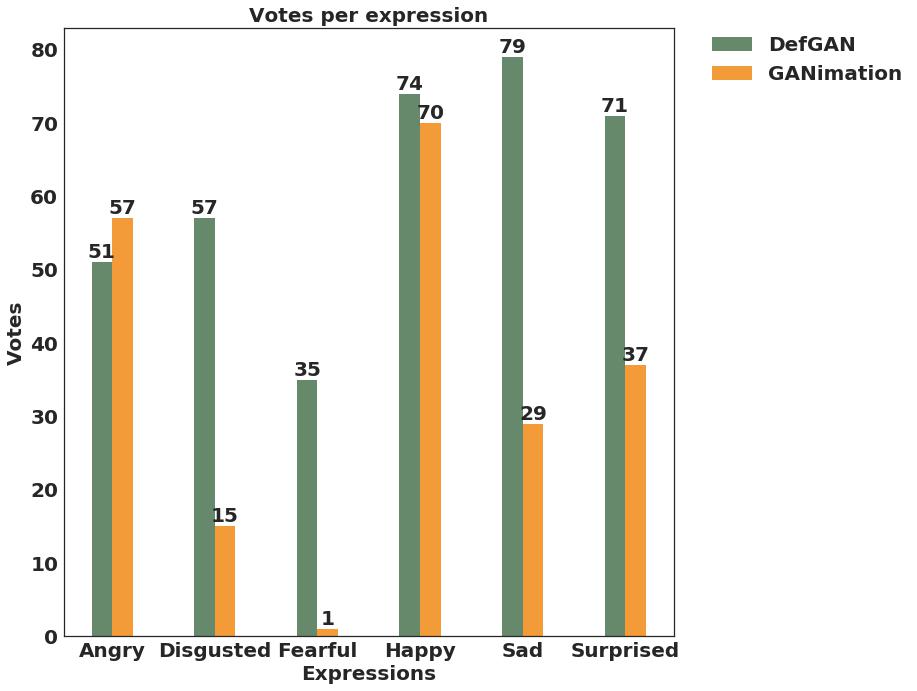}
    \end{tabular}
    \caption{\small{Results of the User Study. Here we show the results of the second stage of the user study. About 63\% of the total votes went to DefGAN as compared to 37\% to GANimation \cite{pumarola2018ganimation}. Across expressions, the editing results of DefGAN were preferred over the editing results of GANimaton \cite{pumarola2018ganimation} with the only exception being `Anger' and `Happy' where the results were close.}}
    \label{fig:UserStudy}
\end{figure}
\vspace{-1pt}
We evaluate the quality of expression edits done by DefGAN and GANimation \cite{pumarola2018ganimation} by conducting a user study. We carry out the user study in two stages, in the first stage (55 users) we evaluate how realistic are the editing results of each method,  without directly comparing them. We randomly sample 10 images edited by each method and show them to the users, asking them to give rate the plausibility of the image from 1 (Definitely implausible) to 4 (Definitely plausible). Each user was shown the same set of 20 images (10 from each method, in random order). The results of the user study showed that around \(60\%\)  of all the images shown were rated as plausible with the average plausible image having a score of 3.5 and the average implausible image having a score 1.67. In the next stage of the user study we directly compared the results of DefGAN and GANimation \cite{pumarola2018ganimation}. Sixteen random in-the-wild images were chosen for this stage. These images were selected to have a close to neutral input expression and to not have extreme poses. Each image was assigned a random target expression from the following expressions: happy, disgust, sad, fear, angry, and surprise. The results of our method and GANimation \cite{pumarola2018ganimation} were placed side-by-side and users were asked to judge the quality of each edited image with respect to its fidelity to the facial identity of the input image, the closeness to the target expression and the overall plausibility of the image. The order in which the options were shown was randomly chosen for each question to avoid user  bias. The results of the user study are shown in \fig{UserStudy}. Users mostly preferred the results of DefGAN over GANimation \cite{pumarola2018ganimation} on the whole and across most expressions. The results were quite close when the target expression was `angry' and `happy' but were overwhelmingly in our favor for all other expressions. Further details of the user study are given in the appendix.
The results of the user study provide further evidence that expression editing results of DefGAN are not only more realistic but also preserve facial identity and attain the target expression better than GANimation.

\section{Conclusion}

We presented  a novel method for facial expression editing that can produce high quality expression edits on in-the-wild images. We leverage the most recent advances in deformation modelling \cite{shu2018deforming} and expression editing \cite{pumarola2018ganimation} to create a method that is able to learn facial movements as deformations without using ground-truth deformation annotations.
The explicit use of a deformation in the ``motion-editing'' phase allows DefGAN to perform targeted edits on the input face which extensive evaluations show are not only able to produce very high quality edited images but also better retain expression invariant facial attributes. In future work, we hope to improve expression modelling by taking into account also the temporal nature of expression and possibly extending expression editing to in-the-wild video sequences.

\paragraph{Acknowledgements.} We would like to thank Francesc Moreno-Noguer for his valuable comments. This work was supported by a gift from Adobe, NSF grants CNS-1718014 and DMS 1737876, the Partner University Fund, and the SUNY2020 Infrastructure Transportation Security Center.

{\small
\bibliographystyle{ieee_fullname}
\bibliography{refs}
}

\FloatBarrier

\begin{appendix}

\begin{center}\LARGE\bfseries
Appendix
\end{center}

\section{Additional Results on Expression Transformations}

\begin{figure*}[h!]
    \centering
    \includegraphics[width=0.95\textwidth]{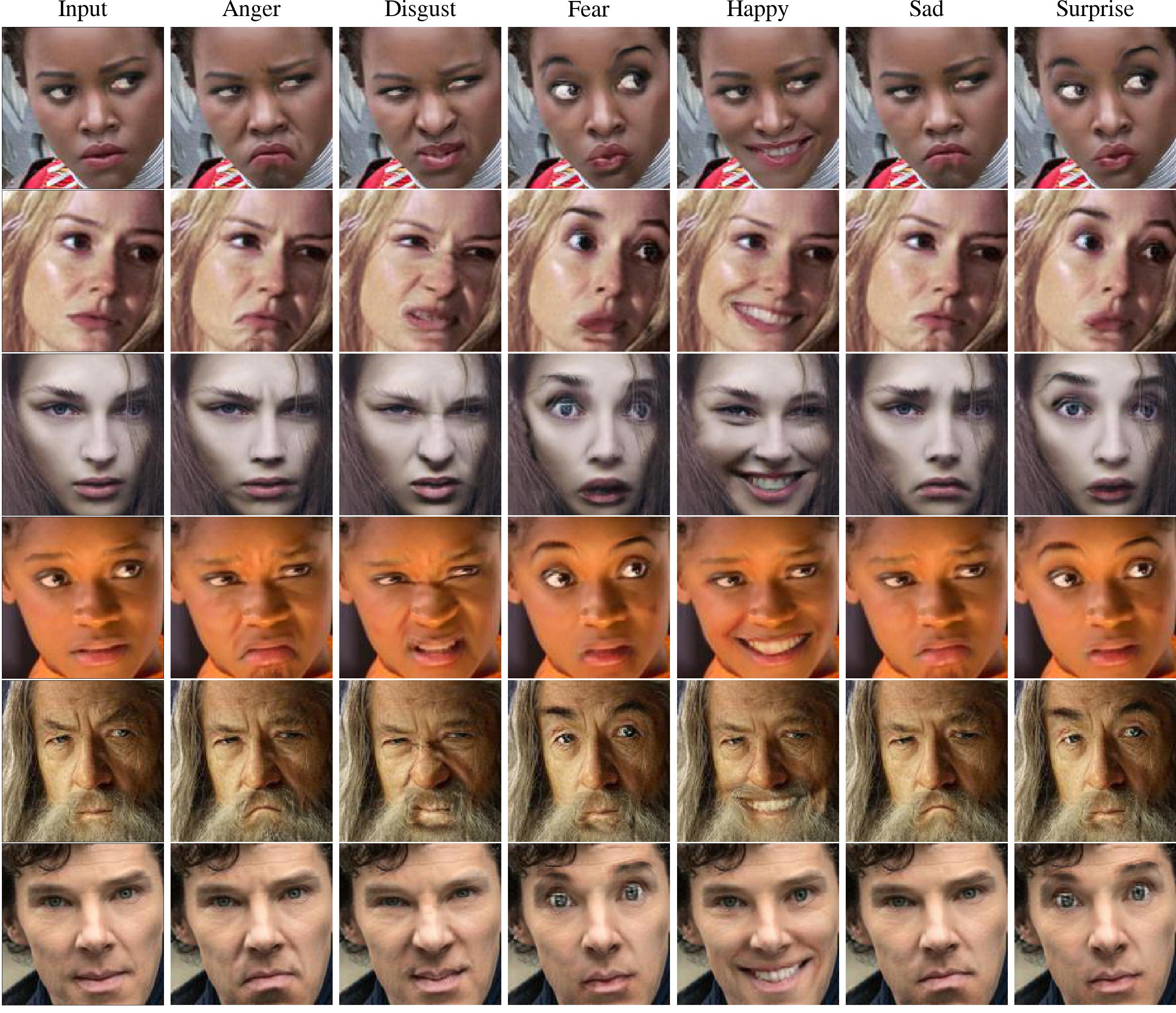}
    \caption{\textbf{Expression Transformation with DefGAN.} We provide additional results on expression transformation using our approach. Our results faithfully capture the details of each facial expression for each example, while successfully maintaining characteristic details of the person in the input image.}
    \label{fig:CelebAHQ_suppmat}
\end{figure*}

In this section we provide additional results of expression transformations as well as comparisons with previous state-of-the-art \cite{pumarola2018ganimation}. 
In \fig{CelebAHQ_suppmat}, we show expression transformations using DefGAN on face images with variations in pose, lighting and ethnicity. Our results faithfully capture the details of each facial expression in each example while successfully maintaining characteristic details of the person in the input. Thanks to deformation disentanglement, the deformation generator successfully captures the facial movement of each expression, generating vivid details on facial regions such as the eyebrows, mouth etc.

\begin{figure}[ht]
    \centering
    \includegraphics[width=.49\linewidth]{figures/Histograms/FaceID/FaceDist_noerr.jpg}
    \includegraphics[width=.49\linewidth]{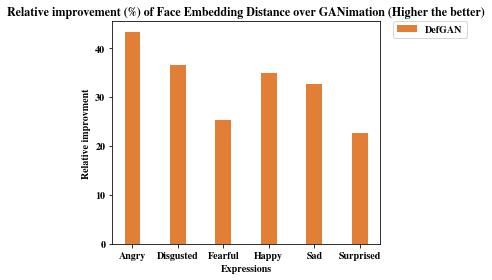}
    \caption{\textbf{Face Embedding Distance.} \textbf{Left: } Here we show the distance between the CMU-OpenFace \cite{amos2016openface} embeddings of the input image and the images edited by DefGAN and GANimation. As one can see, DefGAN's edits consistently preserve facial identity better than GANimations's. \textbf{Right: } Here we show the average relative improvement of the Face Embedding Distance between DefGAN's and GANimation's edits. We see that ``Angry'' has the highest relative improvement.}
    \label{fig:hists_suppmat}
\end{figure}

\begin{figure*}[ht]
    \centering
    \hspace{-5pt}
    \includegraphics[width=1.0\textwidth]{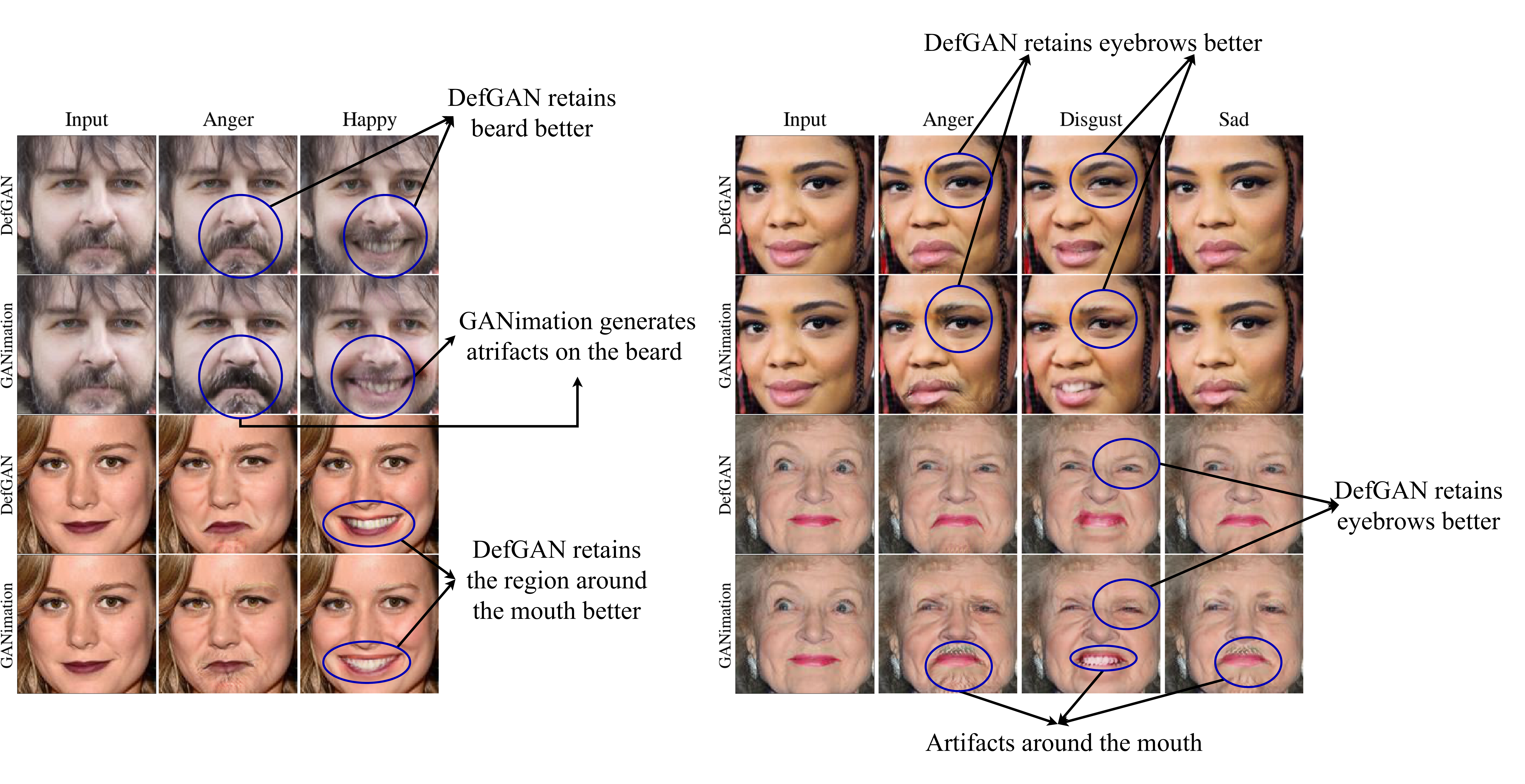}
    \caption{\textbf{Result comparison of DefGAN and GANimation~\cite{pumarola2018ganimation} in details.} We hereby demonstrate the clear benefit of DefGAN comparing to GANimation by highlighting details of several expression edits. Due to the explicit deformation modeling for AUs, DefGAN disentangles the editing task into an image deformation followed by a texture hallucination. This allows each generator within DefGAN, i.e the deformation generator and the texture generator, to perform a more focused task. As a result, our expression editing results better retain characteristic details of the input face, such as facial hair, lip color and eyebrows. On the contrary, GANimation~\cite{pumarola2018ganimation} sometimes produces seemingly arbitrary details, including missing/inaccurate facial hair, missing eyebrows etc.
    }
    \label{fig:DefGANvGAN_suppmat}
\end{figure*}

In \fig{DefGANvGAN_suppmat}, we highlight the utility of explicit  deformation modeling for facial expressions by providing some detailed comparisons with the prior state-of-the-art, GANimation~\cite{pumarola2018ganimation}.  Due to the explicit deformation modeling for AUs, DefGAN disentangles the editing task into an image deformation followed by a texture hallucination. This allows each generator within DefGAN, i.e the deformation generator and the texture generator, to perform a more focused task. As a result, our expression editing results better retain characteristic details of the input face, such as facial hair, lip color and eyebrows. In contrast, GANimation~\cite{pumarola2018ganimation}, sometimes produces seemingly arbitrary and unwanted facial details, such as missing/inaccurate facial hair or missing eyebrows. 

A consequence of DefGAN better preserving characteristic facial details of the input is that it also preserves facial identity better. In \fig{hists_suppmat}, we provide further numerical evidence of this by comparing the CMU-OpenFace\cite{amos2016openface} embeddings of the edited image and the input image using GANimation \cite{pumarola2018ganimation} as a baseline. CMU-OpenFace \cite{amos2016openface} uses a facial recognition network to map an input face image to a 128-dimensional embedding. The distance between two facial identities in this embedded space is given by \(1 - cos(I_{x}, I_{y})\) where \(I_{x} \text{ and } I_{y}\) are the embeddings and \(cos\) is the cosine similarity. We calculate this facial embedding distance for fifty different instances of AUs for each expression. More specifically, we randomly select fifty different people from the RaFD Dataset \cite{langner2010presentation} and calculate the AUs using Cambridge-Openface \cite{baltrusaitis2018openface} on their frontal expression-annotated images. We see that across all expressions and their instances DefGAN preserves facial identity better than GANimation \cite{pumarola2018ganimation}. 

\section{Additional Results on AU Activations}
In this section, we complement the results discussed in Section 4.3 of the paper and show the results of manipulating individual AUs on both the deformed image and the output. 

\begin{figure*}[ht]
    \centering
    \hspace{-5pt}
    \includegraphics[width=0.75\textwidth]{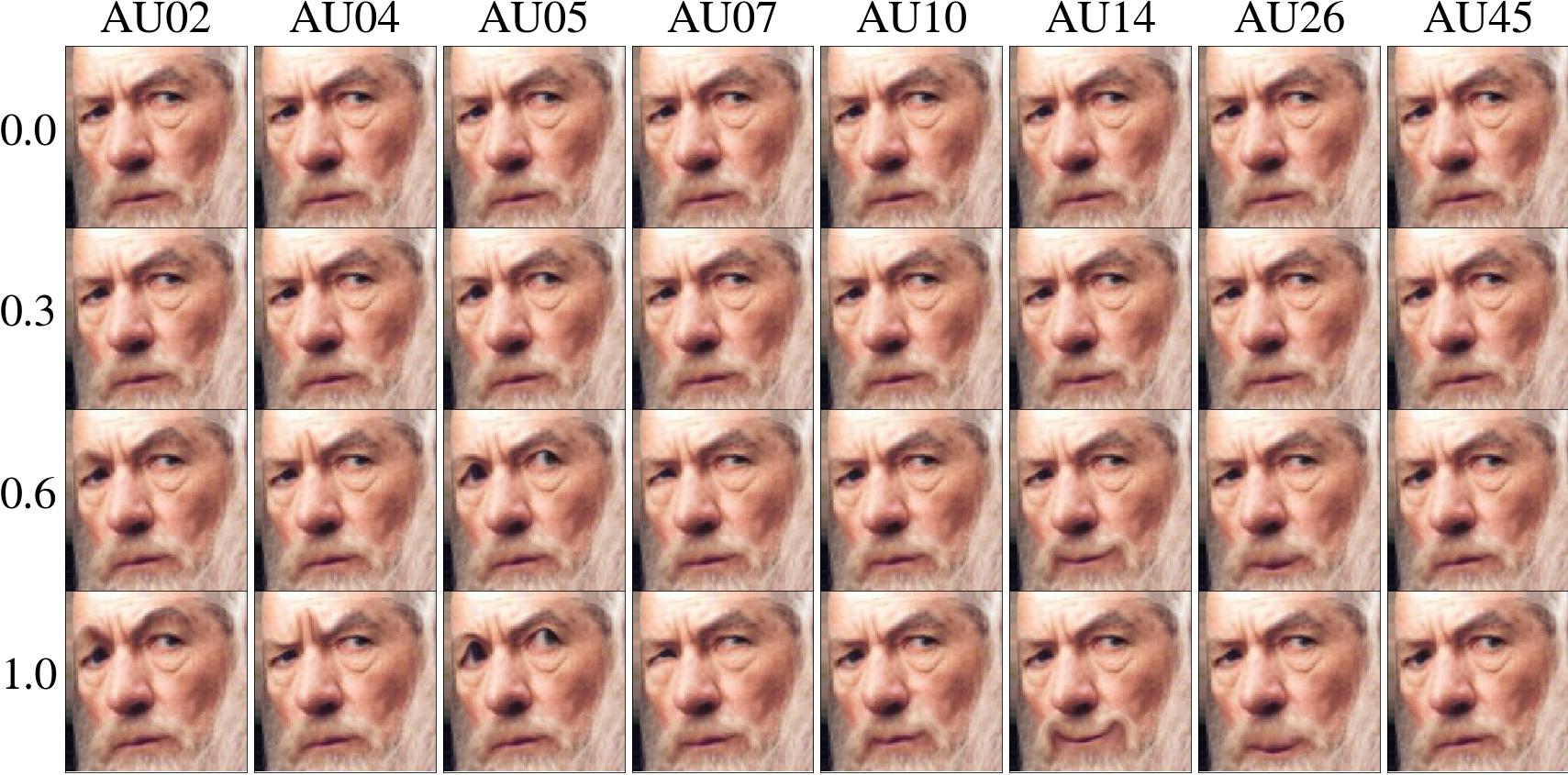}\\
    \includegraphics[width=0.75\textwidth]{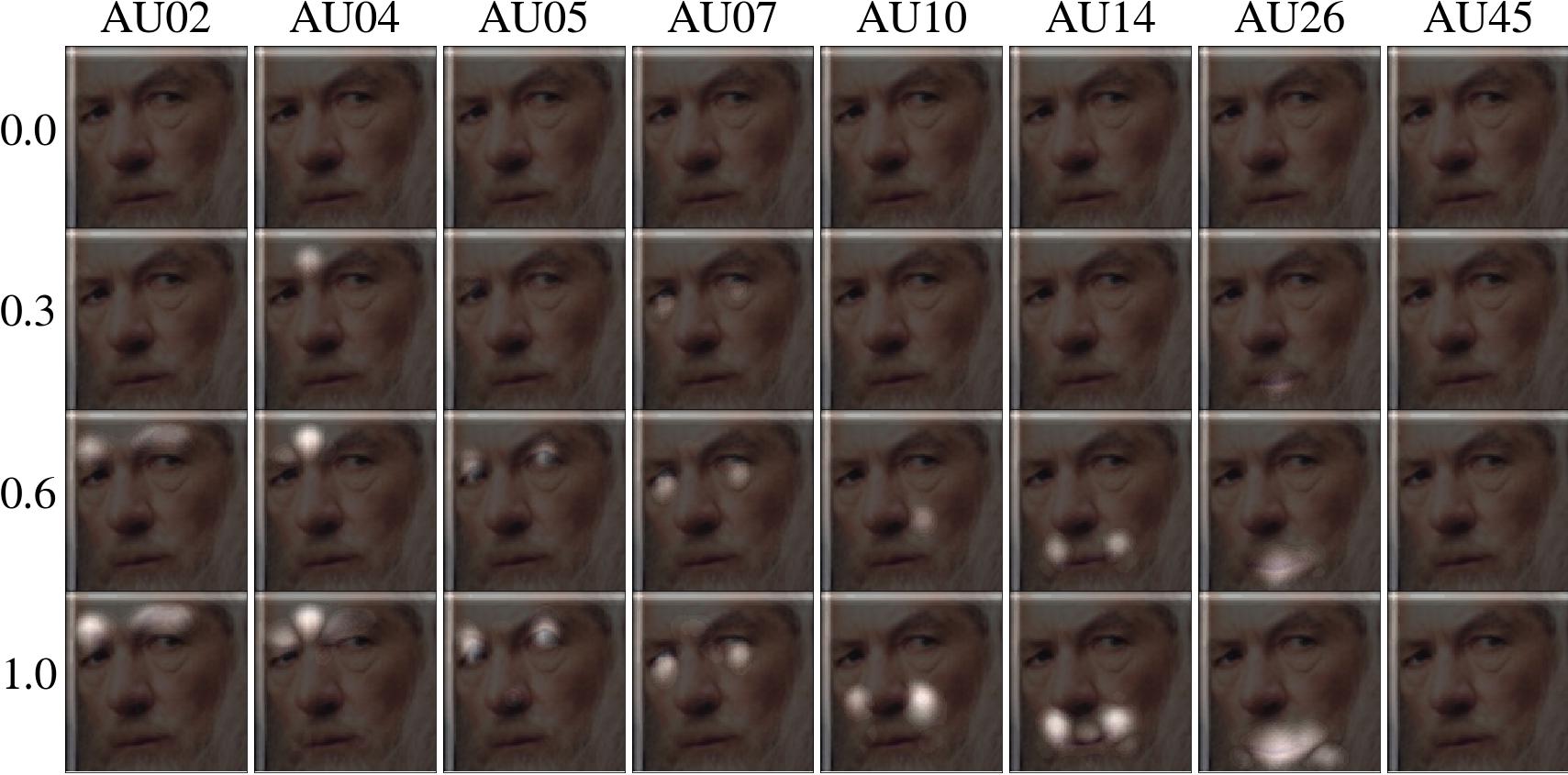}
    \includegraphics[width=0.75\textwidth]{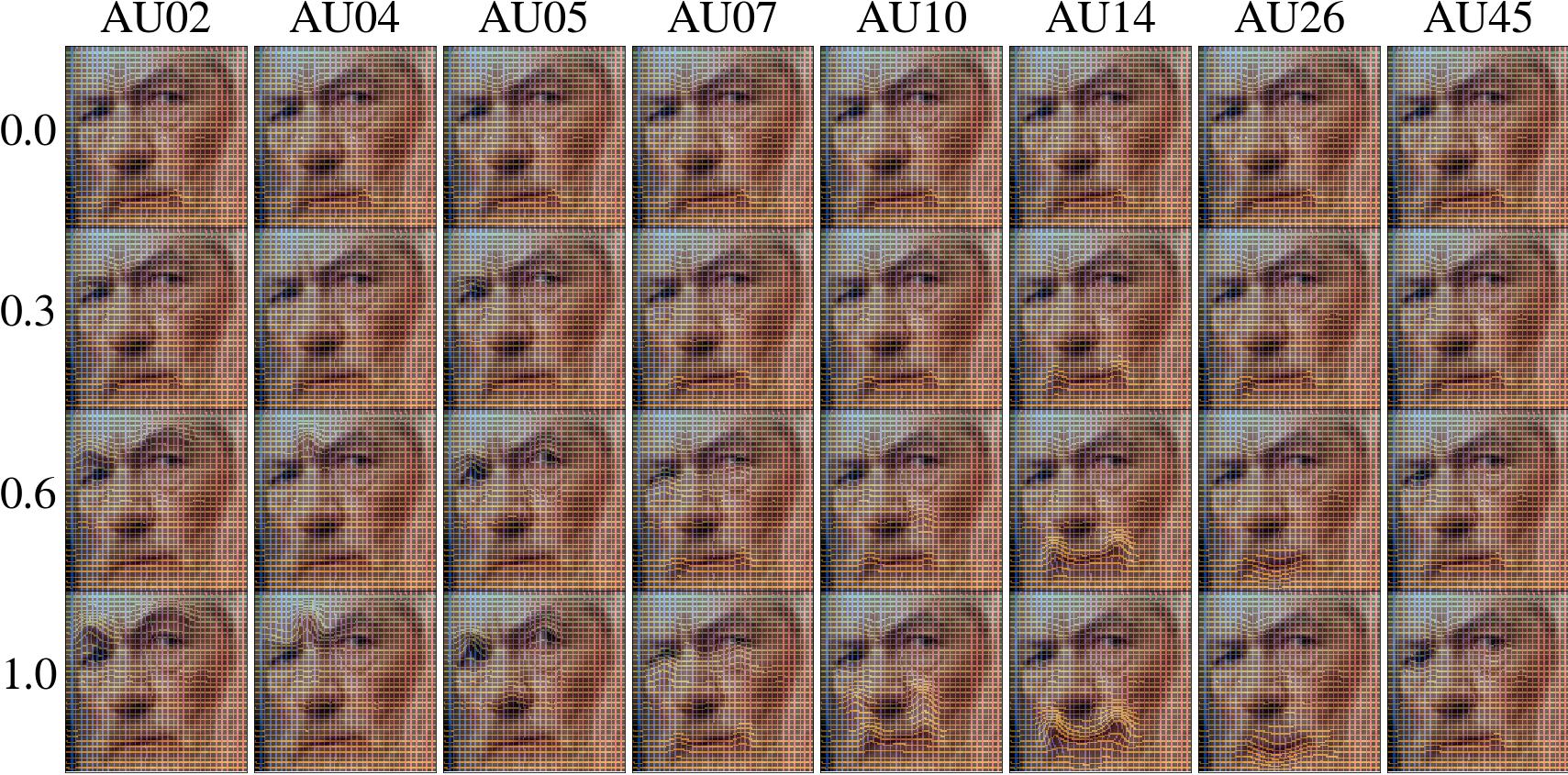}
    \caption{\textbf{Effects of Single AU editing on the Deformed image} \(\cal{I}_{\b{y}}^{*}\). In this figure we show the effects of changing single AU activations on the deformed input image. \textbf{Top: } As can be seen, changing the intensity of AUs causes smooth changes in facial regions. \textbf{Middle: } This figure shows the regions of the face that are deformed as we change the intensity of each AU. We see that the movement is only restricted to regions of the face relevant to the corresponding AU being changed, akin to the results of true facial movement.
    \textbf{Bottom: } Similar to the figure in the middle, looking at the deformation grid, we see that the movement is only restricted to regions of the face relevant to the corresponding AU being changed.}
    \label{fig:AUIntDef_suppmat}
\end{figure*}

In \fig{AUIntDef_suppmat}, we observe the effects of changing the AU intensity on the deformed image \(\cal{I}_{\b{y}}^{*}\). We clearly see that DefGAN is able to faithfully model facial movements through its deformation mechanism. For example, with the increasing intensity of AU02 (Outer Brow Raiser) and AU07 (Lid Tightener) we see that the eyebrows moving up and we see the eyes becoming smaller respectively. Similarly, with increasing intensity of AU26 (Jaw Drop) and AU14 (Dimpler) we see changes around the mouth.

\begin{figure*}[ht]
    \centering
    \includegraphics[width=0.75\textwidth]{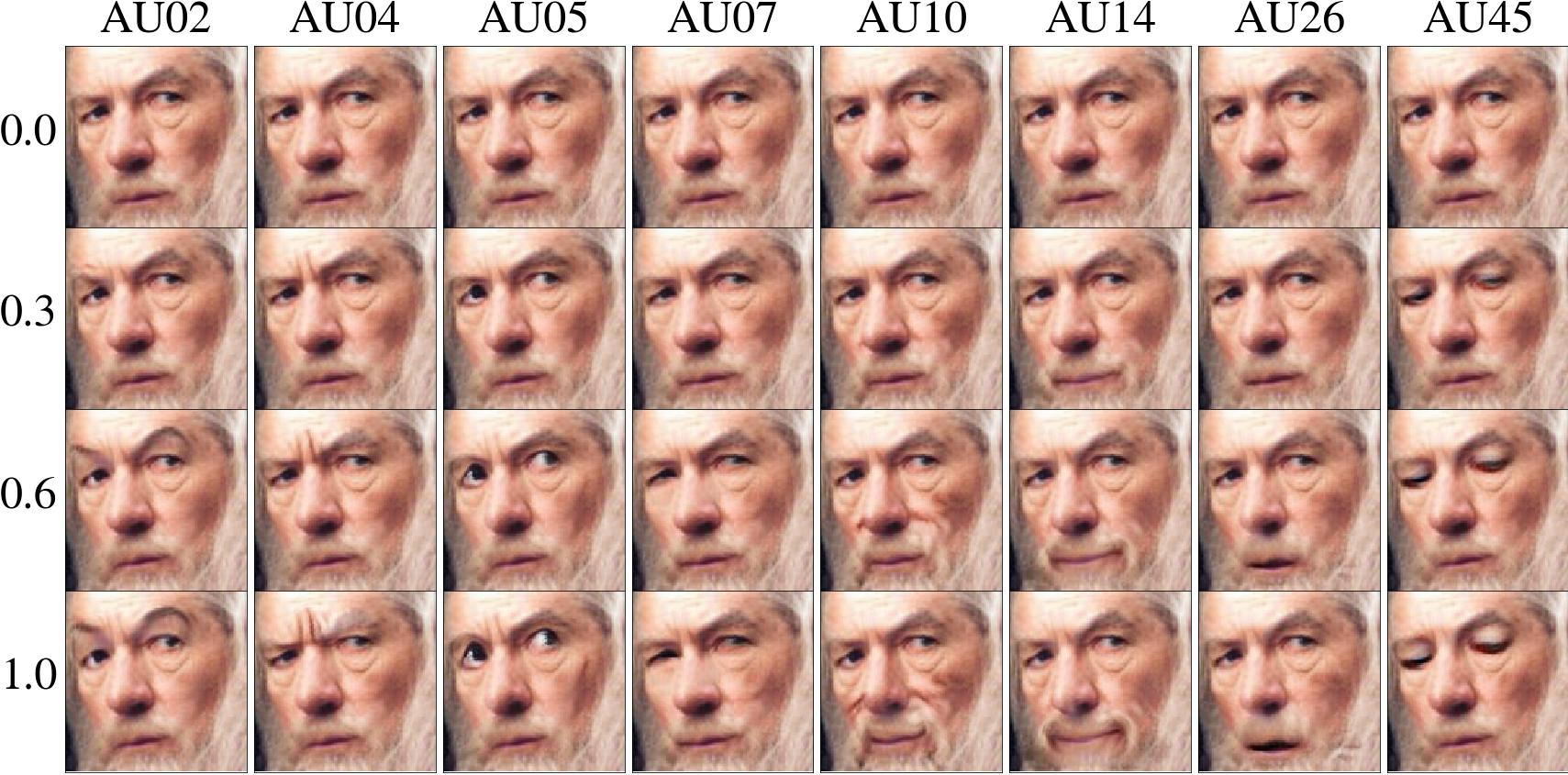}
    \caption{\textbf{Effects of Single AU editing on the Output Image} \(\cal{I}_{\b{y}}\). In this figure we show the effects of single AU activations on the final edited image. As can be seen, changing the intensity of AUs causes smooth changes facial expression without any artifacts or inaccurate texture hallucinations.}
    \label{fig:AUIntOut_suppmat}
\end{figure*}

AU26, which represents a ``Jaw Drop" is a good example of an AU who's intensity change entails both facial movement and the generation of new textures (the opening of the mouth), and we see that DefGAN handles both really well. In the ``AU26" column of \fig{AUIntDef_suppmat} we see DefGAN changing the region around the mouth using deformations (mimicking facial muscle movements) and in \fig{AUIntOut_suppmat} (``AU26" column) we see DefGAN hallucinating the texture of an open mouth thus completing the editing process. 

In cases like that of AU45 (Blink), which requires significant generation of new textures (the opening and closing of eyelids), we see that DefGAN producing little change in the deformed image (``AU45'' column of \fig{AUIntDef_suppmat}) and generating the texture of the eyelids in the final output (``AU45'' column of \fig{AUIntOut_suppmat}).

\section{Additional Results for Concentration of Edits}
\begin{figure*}[ht]
    \centering
    \includegraphics[width=0.8\textwidth]{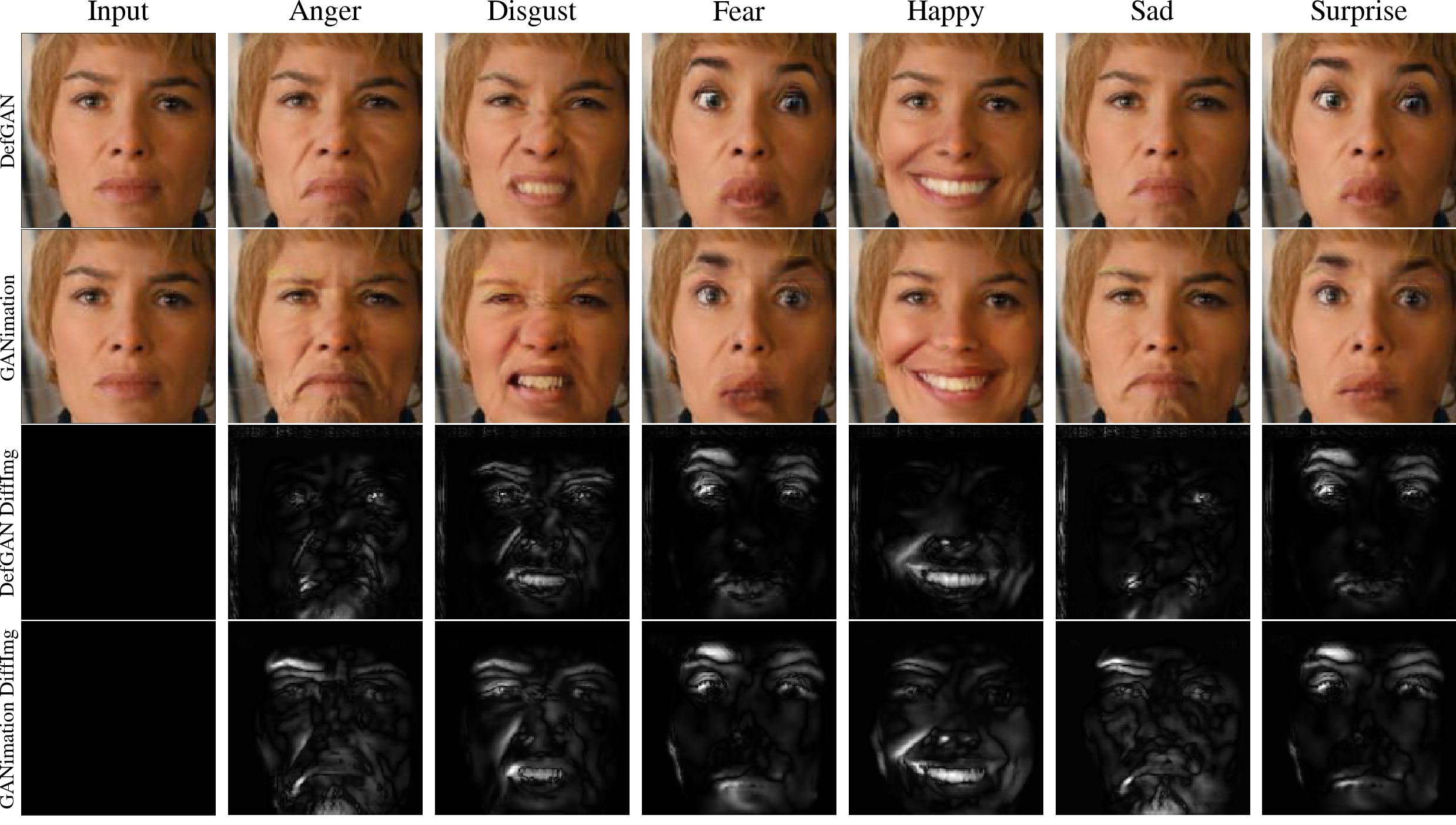}\\
    \includegraphics[width=0.8\textwidth]{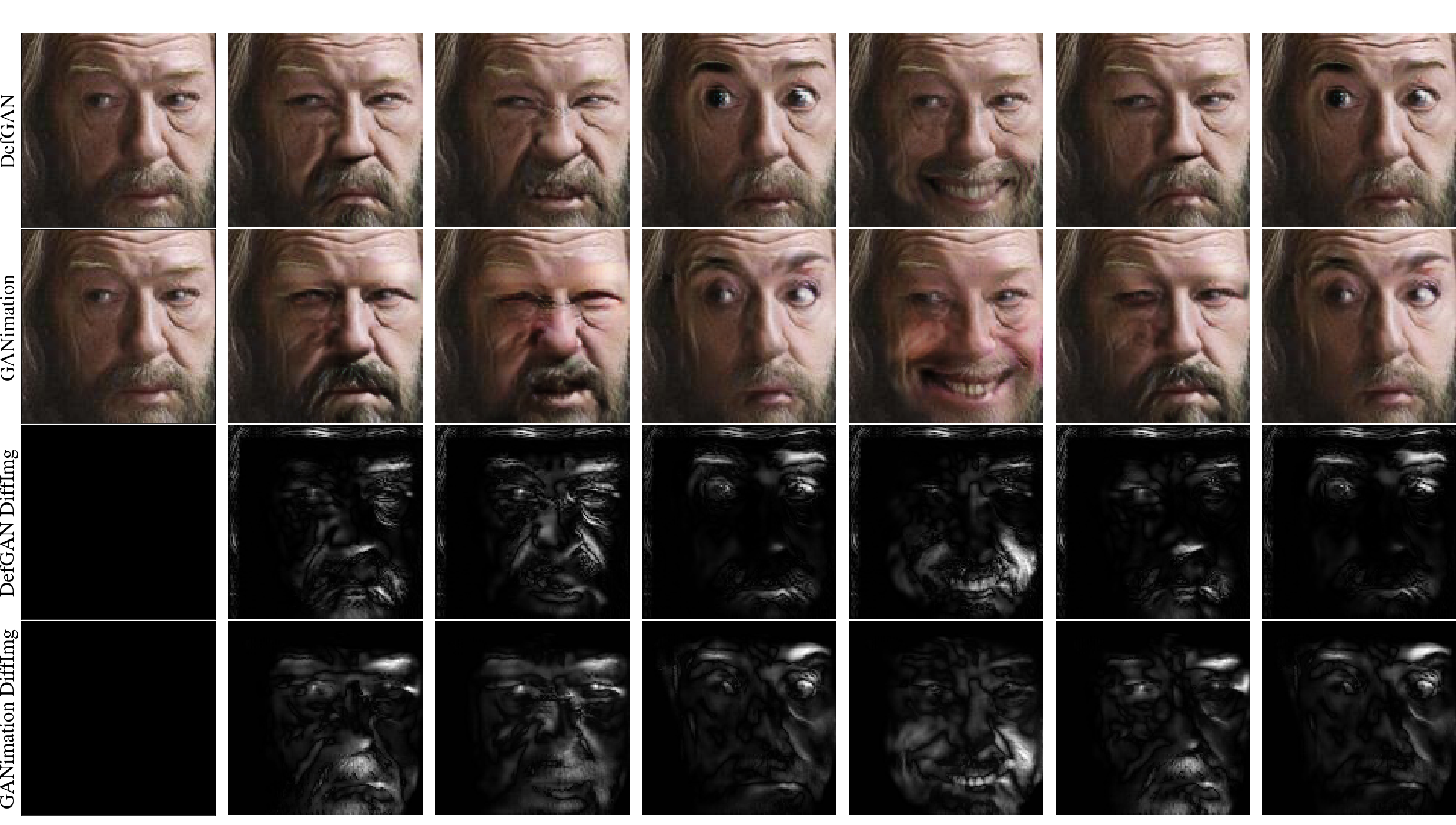}\\
    \caption{\textbf{Difference Images.} The rows marked as ``GANimation DiffImg" and ``DefGAN DiffImg" show the absolute pixel-wise difference between the edit made by the respective methods and the input image. As can be seen, DefGAN's edits are more concentrated to the region relevant to the expression transformation. For example, the transformation to `Happy' and `Sad' of both people shown above. DefGAN almost exclusively changes the area around the mouth while GANimation's edits are spread out all across the input face.}
    \label{fig:Diff1_suppmat}
\end{figure*}

\begin{figure*}[ht]
    \centering
    \includegraphics[width=0.8\textwidth]{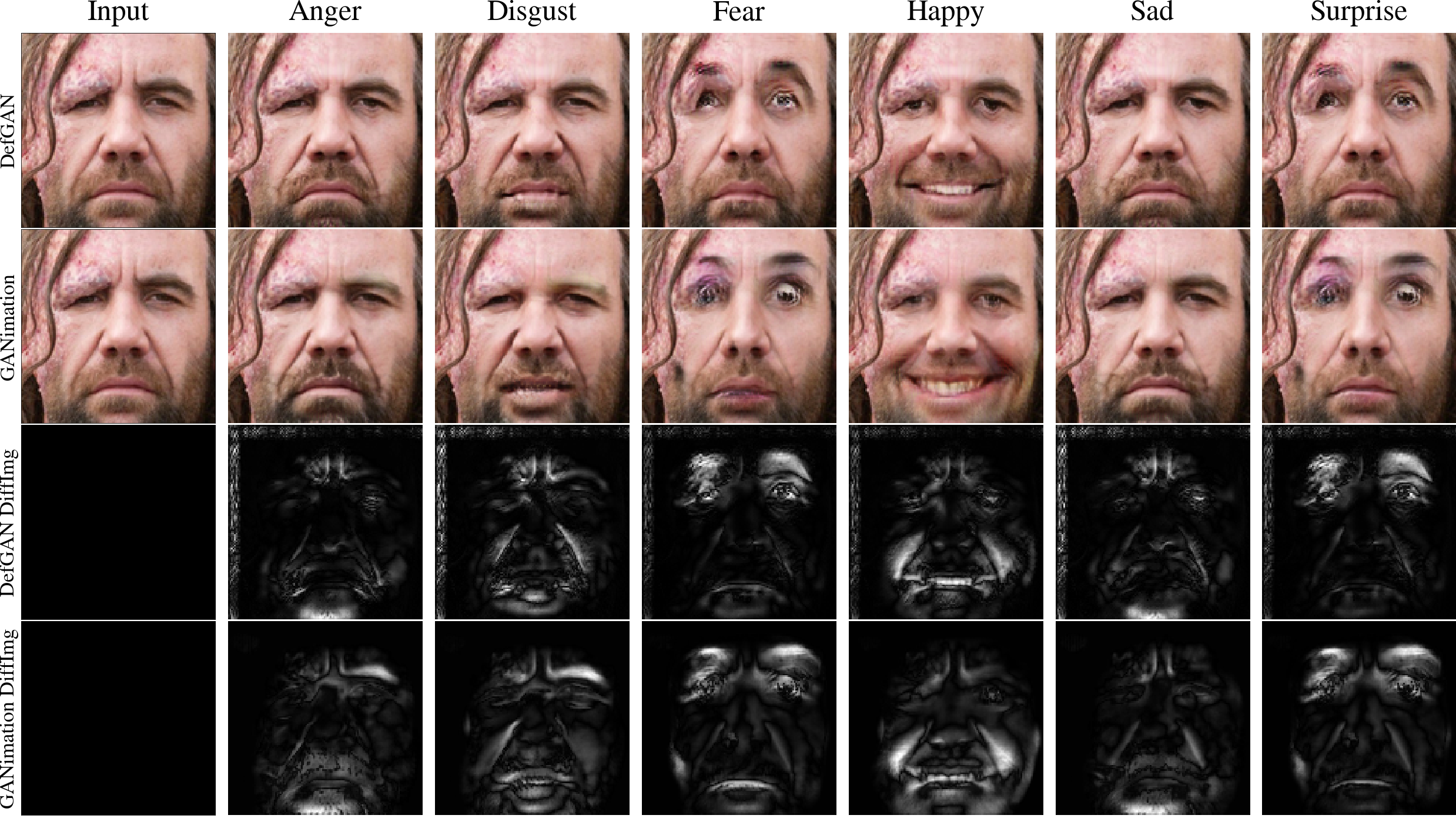}\\
    \includegraphics[width=0.8\textwidth]{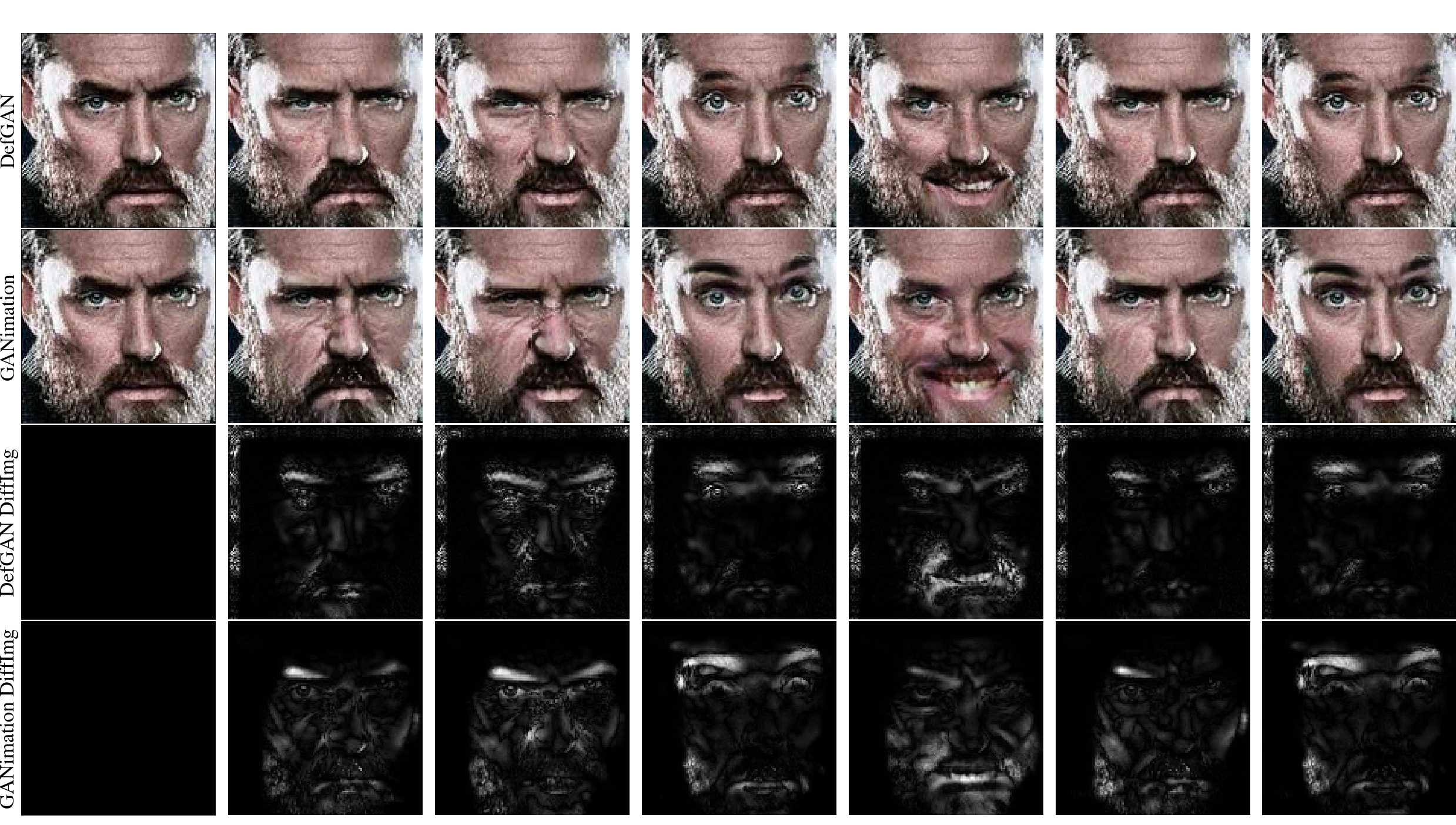}
    \caption{\textbf{Difference Images.} The rows marked as ``GANimation DiffImg" and ``DefGAN DiffImg" show the absolute pixelwise difference between the edit made by the respective methods and the input image. As can be seen, DefGAN's edits are more concentrated to the region relevant to the expression transformation. For example, the transformation to `Anger' and `Fear' of both people shown above. DefGAN almost exclusively changes the area around the eyebrows and the mouth GANimation's edits are spread out all across the input face.}
    \label{fig:Diff2_suppmat}
\end{figure*}

In this section, we provide further evidence (in addition to that given in Section 4.2 of the paper) of DefGAN's edits being more targeted and concentrated than GANimation's, using the difference image. \fig{Diff1_suppmat} and \fig{Diff2_suppmat} show the absolute pixel-wise difference between the images edited by DefGAN and GANimation and the input image. 

In \fig{Diff1_suppmat}, the biggest differences can be seen when the target expressions are ``Disgust'', ``Happy'' and ``Sad''. Specifically, in the case of the person at the bottom of \fig{Diff1_suppmat} (the old man with a beard) we see than GANimation makes significant changes to the beard and the eyebrows when the target expressions are ``Disgust'' and ``Sad''.

Similarly, in \fig{Diff2_suppmat} we see the biggest changes in ``Anger" and ``Fear". 
We also observe the same effects in \fig{DefGANvGAN_suppmat} where GANimation produces more change than desired. This tendency of GANimation to ``overedit" also manifests itself in \fig{hists_suppmat} where we clearly see that it introduces more change in facial identity of the edited image than DefGAN does.

\FloatBarrier
\section{Training Details}
\subsection{Architecture Details}
\paragraph{The texture generator \(G_{\text{Texture}}\).} The texture network is identical to the one used in GANimation \cite{pumarola2018ganimation} which builds upon the variation of the network proposed by Johnson et al. \cite{johnson2016perceptual} and was used by Zhu et al. in \cite{CycleGAN2017} to achieve impressive results for image-to-image mapping.
\paragraph{The deformation generator \(G_{\text{Def}}^{W}\).} The deformation generator is identical to the texture generator except that we replace the last three convolutional layers (the layers responsible for upsampling) with Bilinear Upsampling layers.
\paragraph{The Discriminator \(D\).} The Discriminator has a PatchGAN \cite{pix2pix2016} architecture where each element of the output matrix \(X_{ij}\) represents the probability of the overlapping patch \(ij\) to be real. We also add an AU output head to the penultimate layer of \(D\) that estimates the AU output \(\b{x} = (x_{1},\hdots, x_{N})\) of the input image \(\cal{I}_{\b{x}}\).

\subsection{Coefficient Values}
During training we use the following coefficient values for the losses
\begin{align}
    \lambda_{exp}^{G_{Comp}} = 4000.0\\
    \lambda_{exp}^{G_{Def}} = 1000.0\\
    \lambda_{cyc} = 100.0\\
    \lambda_{comp} = 10.0\\
    \lambda_{eye}^{\cal{G}} = 0.1\\
    \lambda_{TV}^{\cal{G}} = 1e-5\\
    \lambda_{eye}^{\cal{M}} = 0.1\\
    \lambda_{TV}^{\cal{M}} = 1e-5\\
\end{align}

\section{User Study}
\begin{figure}[h!]
    \centering
    \includegraphics[width=0.6\linewidth]{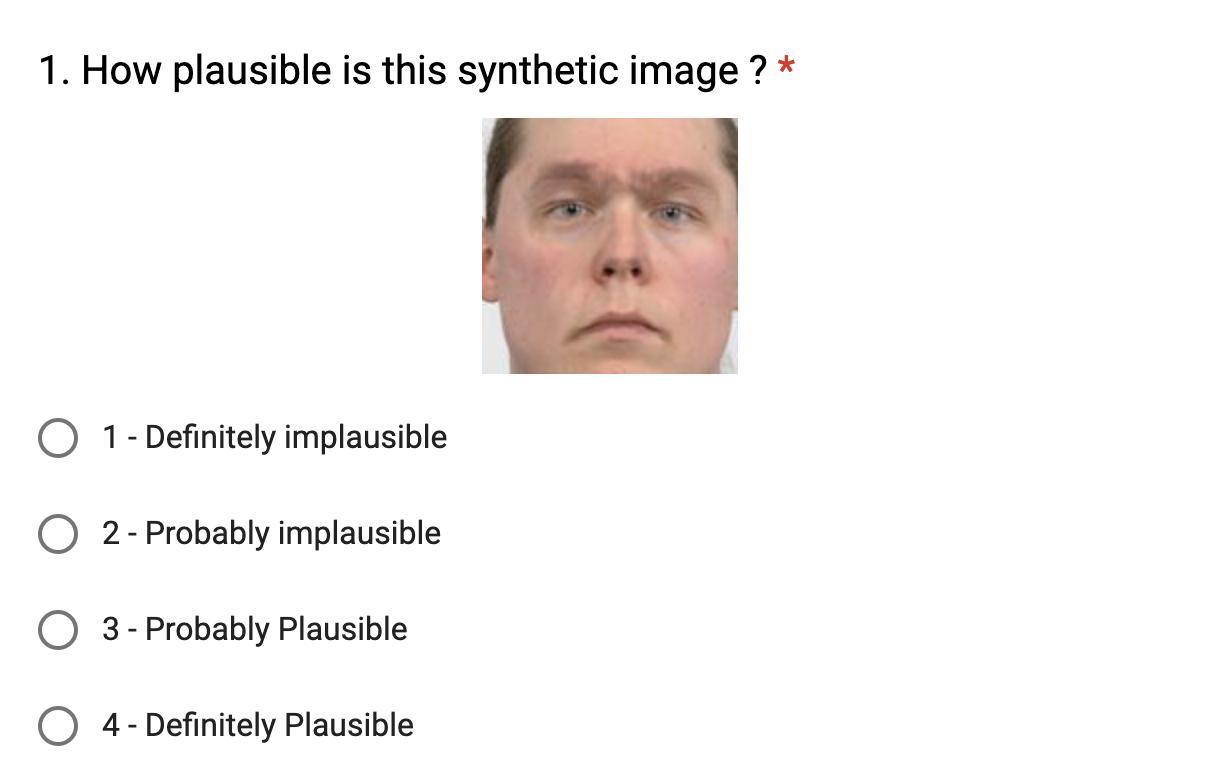}
    \caption{\textbf{User Study Stage 1. }Here we show an example of a question asked in the first stage of the user study.}
    \label{fig:UserStudy1_suppmat}
\end{figure}
\begin{figure}[h!]
    \centering
    \includegraphics[width=0.6\linewidth]{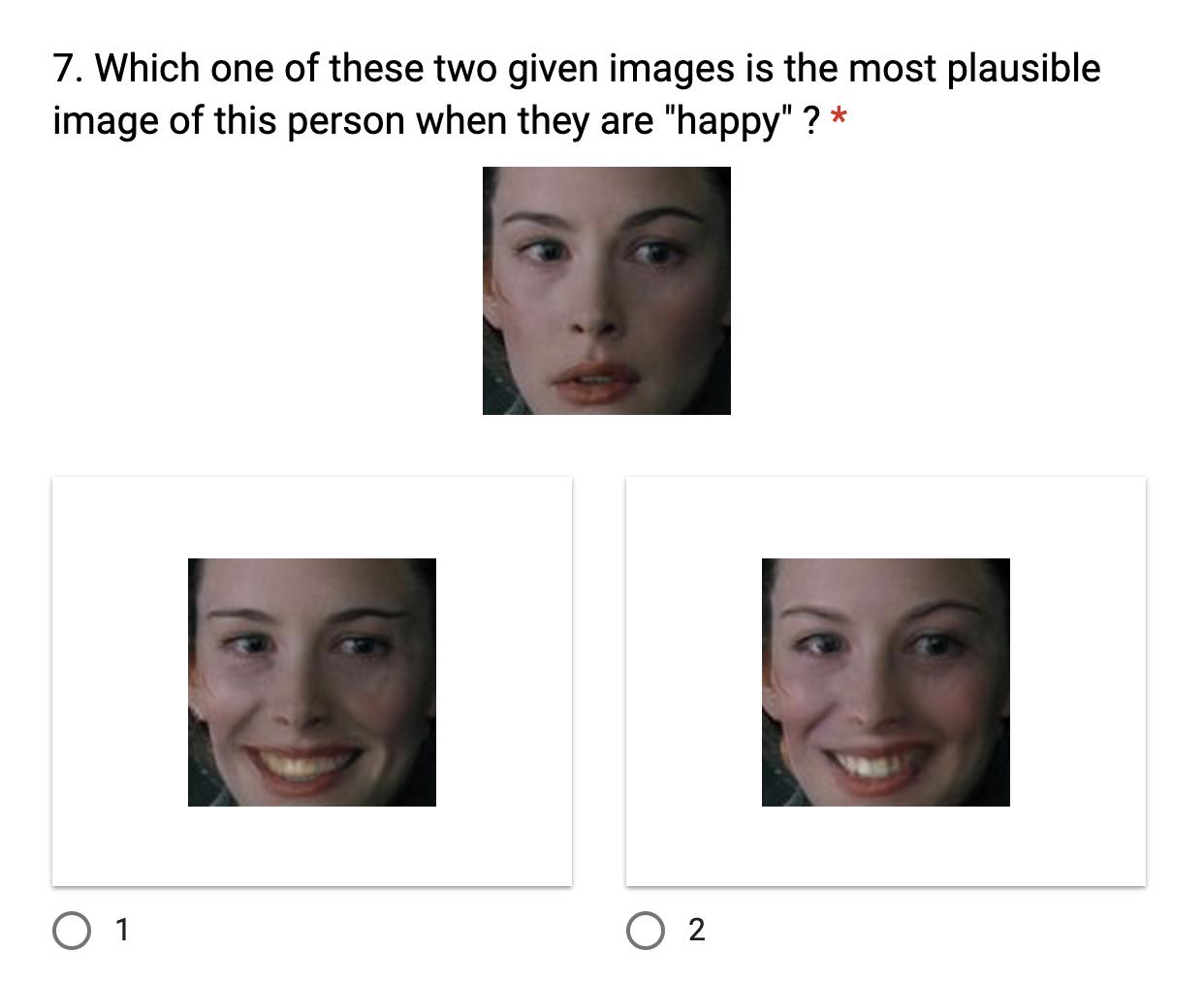}
    \caption{\small{\textbf{User Study Stage 2. }Here we show an example of a question asked in the second stage of the user study.}}
    \label{fig:UserStudy2_suppmat}
\end{figure}

We evaluate the realism of edits made by DefGAN and GANimation \cite{pumarola2018ganimation} independently and also perform a direct comparison of their edits using user studies. In the first stage, we ask users to rate the `Plausibility' of an image given in the question, where the image shown is an edit made either by DefGAN or GANimation \cite{pumarola2018ganimation}, an example form is given in \fig{UserStudy1_suppmat}. 

In the next stage, we perform a direct comparison between the images edited by GANimation \cite{pumarola2018ganimation} and DefGAN and we ask the user to choose the image that is more `plausible' and is also more faithful to the target expression, an example of the form is given in \fig{UserStudy2_suppmat}. The order of the results shown was randomly chosen for each question.

\section{Comparison with StarGAN}
\begin{figure*}[h!]
    \centering
    \includegraphics[width=0.7\textwidth]{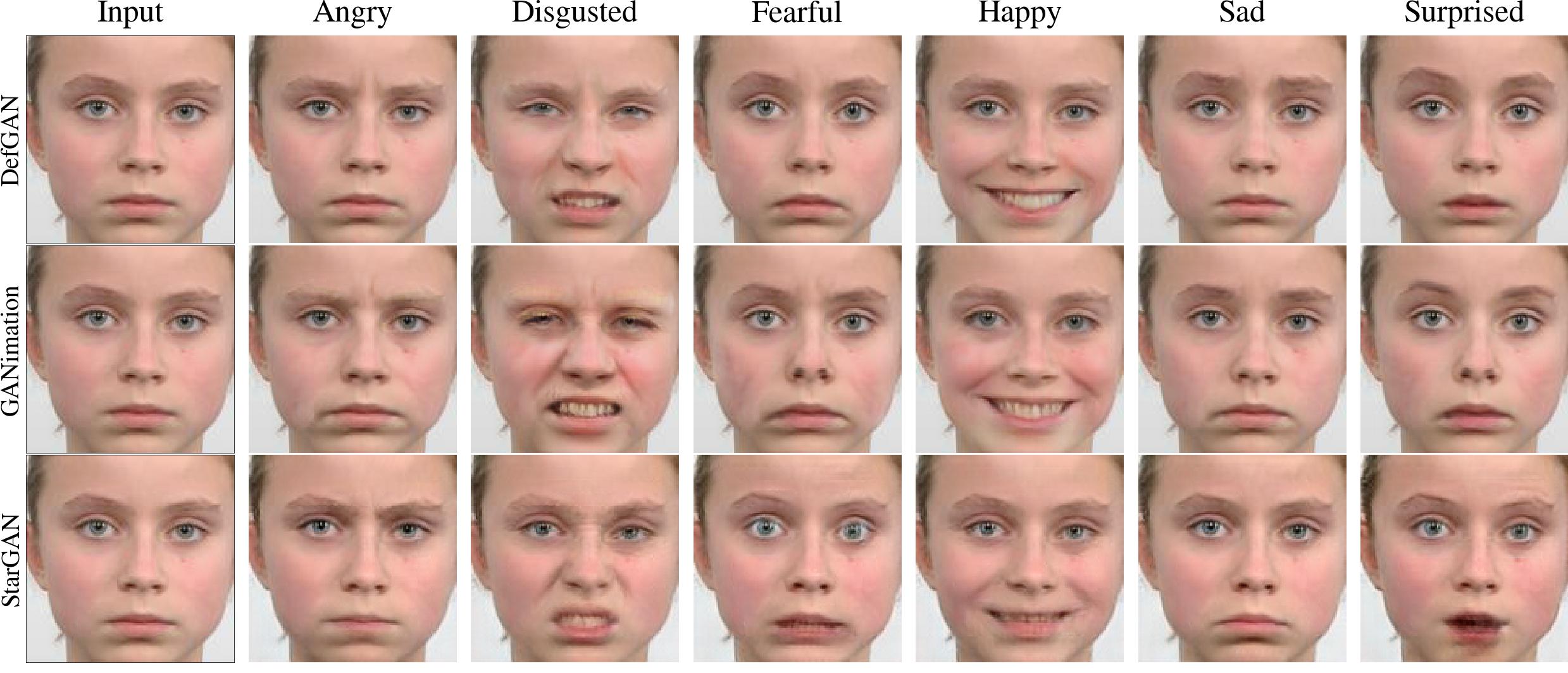}
    \caption{\textbf{StarGAN on RaFD. }In this figure we compare DefGAN, GANimation \cite{pumarola2018ganimation} and StarGAN \cite{StarGAN2018} on the RaFD Dataset \cite{langner2010presentation}.}
    \label{fig:StarGANRafd_suppmat}
\end{figure*}

\begin{figure*}[h!]
    \centering
    \includegraphics[width=0.7\textwidth]{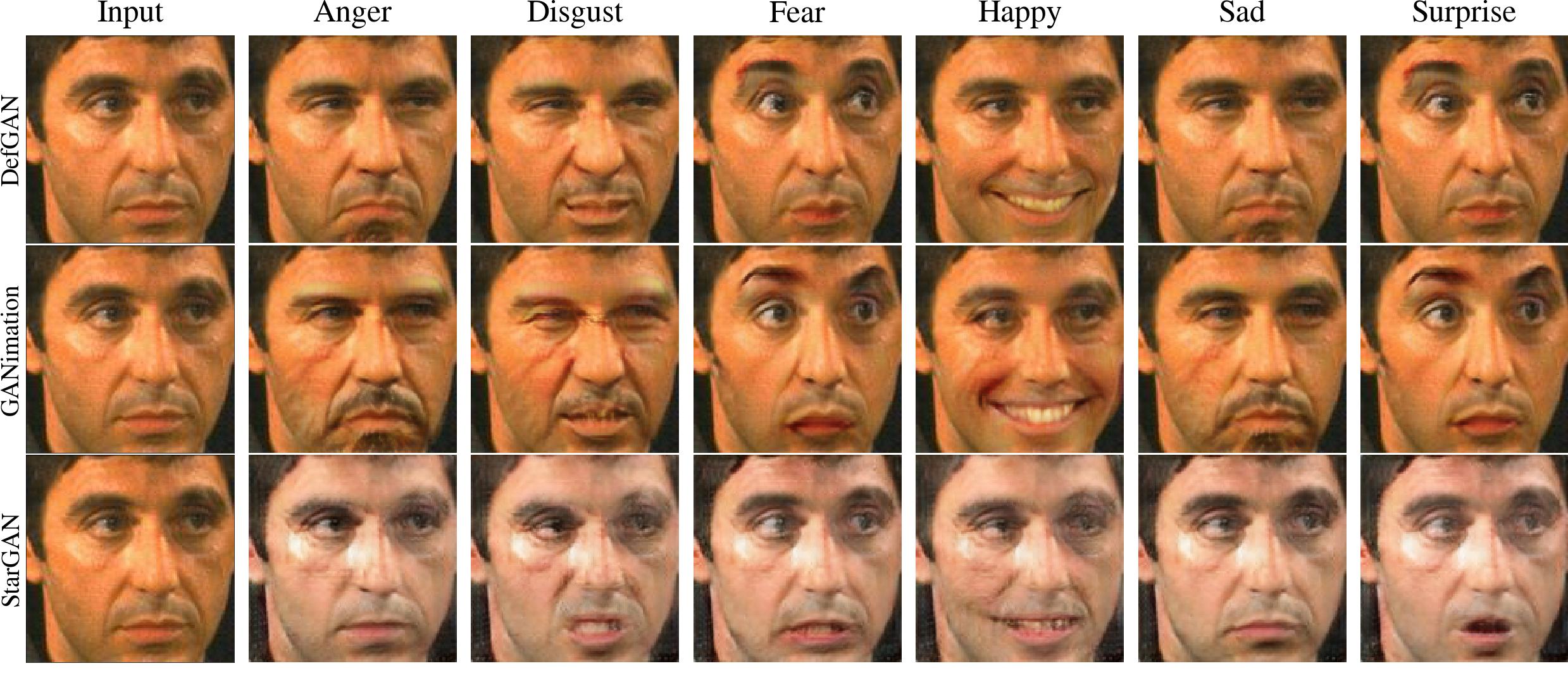}
    \caption{\textbf{StarGAN on in-the-wild. }In this figure we compare DefGAN, GANimation \cite{pumarola2018ganimation} and StarGAN \cite{StarGAN2018} on an in-the-wild image.}
    \label{fig:StarGANWild_suppmat}
\end{figure*}

Here we compare against StarGAN \cite{StarGAN2018} on the RaFD Dataset \cite{langner2010presentation} and on in-the-wild images. In \fig{StarGANRafd_suppmat}, we see that both DefGAN and GANimation perform on par with StarGAN, especially considering StarGAN was trained on RaFD \cite{langner2010presentation} while GANimation \cite{pumarola2018ganimation} and DefGAN were not. On the flip side, when edits are carried out on in-the-wild images as seen in \fig{StarGANWild_suppmat} StarGAN performs much worse for the same reason.

\end{appendix}

\end{document}